\documentclass[letterpaper]{article}
\usepackage{aaai2027}
\nocopyright
\usepackage[hyphens]{url}
\usepackage{graphicx}
\usepackage{natbib}
\usepackage{caption}
\usepackage{algorithm}
\usepackage{algorithmic}
\usepackage{newfloat}
\usepackage{listings}
\DeclareCaptionStyle{ruled}{labelfont=normalfont,labelsep=colon,strut=off}
\floatstyle{ruled}
\newfloat{listing}{tb}{lst}{}
\floatname{listing}{Listing}
\usepackage{booktabs}
\usepackage{placeins}
\usepackage{multirow}
\usepackage{amsmath}
\usepackage{amssymb}
\title{Spatial Supervision Without Attribution Optimization:\\ Improving Post-Hoc Class Activation Maps via Box-Guided Evidence Routing}
\author{
Wenhao Liang, Liangwei Nathan Zheng, Lin Yue\\
Wei Emma Zhang, Mingyu Guo, Weitong Chen
}
\affiliations{
Adelaide University, Adelaide, Australia
}

\begin{document}
\maketitle

\begin{abstract}
Post-hoc class activation maps (CAMs) are a standard tool for inspecting the evidence behind an image classifier's predictions, yet nothing in ordinary training encourages these maps to be spatially appropriate. We study whether inexpensive spatial supervision can improve a classifier's own predicted-class Grad-CAM without ever optimizing an attribution map. Box-Guided Evidence Routing (BGER) trains a lightweight gate on the final feature map under box or mask supervision and routes classification through the gated features, while Grad-CAM is computed separately at the pre-gate representation, so the evaluated map never enters the training objective. With a BCE routing loss, BGER raises MaxBoxAccV2 from $0.584$ to $0.715$ on CUB-200-2011 and from $0.757$ to $0.832$ on Stanford Dogs at comparable accuracy. Matched controls attribute most of the ResNet-50 gain to the spatial supervision reshaping the backbone rather than to routing itself: when classification bypasses the gate, most of the improvement remains, and detaching gradients through the gate leaves the ResNet-50 result nearly unchanged. The same detachment preserves most of the gain in two DenseNet-121 chest X-ray settings but removes the apparent gain on Swin-T, and directly supervising the CAM reaches stronger localization at a larger accuracy cost. Overall, spatial supervision can improve separately evaluated post-hoc CAMs, but both the mechanism and the size of the benefit depend on the architecture and the evaluation setting.
\end{abstract}

\section{Introduction}

Post-hoc class activation maps such as Grad-CAM \citep{selvaraju2017gradcam} are widely used to inspect where a classifier appears to draw its evidence, yet conventional classifiers are never trained to make these maps spatially appropriate. One can supervise a saliency map directly to improve its localization, but that optimizes an attribution proxy from the same family later used to evaluate it. We ask a different question: can spatial supervision applied to an \emph{internal feature route} transfer to the classifier's own predicted-class CAM, so that the post-hoc map concentrates on the target or lesion region \emph{without any attribution map ever entering the training objective}?

\begin{figure*}[t]
    \centering
    \includegraphics[width=\textwidth]{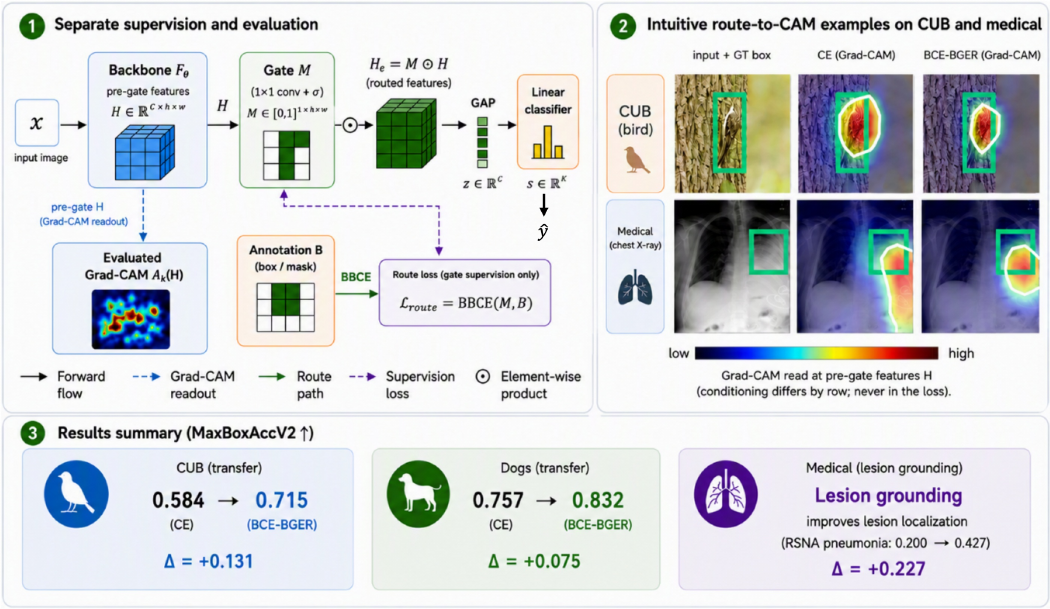}
    \caption{\textbf{Box-Guided Evidence Routing with BCE-BGER.} \textbf{(1)}~The box or mask $B$ supervises only the internal gate $M$ via $\mathcal{L}_{\mathrm{route}}=\mathrm{BBCE}(M,B)$; classification is routed through $H_e=M\odot H$, and the evaluated predicted-class Grad-CAM $A(H)$ is read at the pre-gate features $H$ and never enters the loss. The CAM is not optimized directly, but the class-score gradient still traverses the learned route; backbone-specific detach tests quantify this contribution (Tab.~\ref{tab:arch-probe}; App.~Tab.~\ref{tab:detach-arch},~\ref{tab:detach-densenet}). \textbf{(2)}~Real route-to-CAM examples (seed-0 Grad-CAM at the pre-gate features; GT box green). \emph{CUB}: predicted-class Grad-CAM---CE grounds on bark outside the bird (EIB $0.41$), BCE-BGER on the creeper ($0.65$). \emph{Medical}: RSNA \emph{positive-finding} Grad-CAM lesion grounding on a positive boxed study (conditioned on the abnormality logit, not the predicted class)---CE spreads outside the radiologist box (EIB $0.23$), BCE-BGER concentrates more inside ($0.40$); both classify correctly. The panel's colour-bar label names the shared mechanism (Grad-CAM at pre-gate $H$); the conditioning class differs by row as stated here. To avoid a best case, this is the \emph{median}-improvement study among $491$ eligible positives under a pre-stated rule (App.~\ref{app:rsna-selection}). \textbf{(3)}~Headline MaxBoxAccV2; medical is a controlled extension, not clinical validation.}
\label{fig:method}
\end{figure*}

BGER supervises a minimal gate on the backbone feature map and routes classification through the gated features. The training target is the route; the evaluated map is predicted-class Grad-CAM. Any route-to-CAM improvement must therefore pass through the learned classifier, not through direct optimization of that map. We use \emph{route-to-CAM transfer} to describe transfer under route-supervised training; the matched no-routing control separately measures how much of it routing contributes.

BCE-BGER raises full-test predicted-class Grad-CAM MaxBoxAccV2 from $.584$ to $.715$ on CUB and from $.757$ to $.832$ on Stanford Dogs at similar accuracy. CAM-BCE reaches $.769$ on CUB at a modest accuracy cost. Equal-annotation controls distinguish feature routing from alternative uses of the same boxes. An objective-matched auxiliary control that supervises the identical gate with the identical balanced-BCE loss but classifies from the ungated features reaches $.683$ ($+.099$ over CE), so shared-backbone reshaping explains most of the transfer, while routing adds a smaller $+.032$ increment under the matched ResNet-50 intervention. Lesion-grounding and architecture experiments test where this transfer holds and where it breaks down.

Our contributions are:
\begin{enumerate}
    \item \textbf{BGER} supervises a box/mask-guided feature route and leaves the post-hoc CAM it is evaluated on outside the loss; BCE-BGER uses class-balanced binary cross-entropy for route supervision.
    \item \textbf{Route-to-CAM transfer}: BCE-BGER improves predicted-class CAM localization on CUB and Stanford Dogs, and improves positive-finding lesion maps on two controlled chest-X-ray benchmarks; CAM-BCE provides the direct-supervision comparison.
    \item \textbf{Mechanism analysis.} Matched no-routing and detach interventions show that most of the ResNet-50 transfer arises from changes in the shared representation, with routing adding a smaller increment. The detach experiment reveals strong architecture dependence: the gain survives on ResNet-50 and the reported DenseNet-121 settings but not on Swin-T.
\end{enumerate}

\section{Related Work}

\paragraph{Spatial supervision, explanation regularization, and routing.}
BGER's nearest neighbors fall into two groups that differ in \emph{which object is supervised and which is evaluated}. \emph{Direct explanation-aware approaches} place attribution maps or attribution-derived quantities in the training objective, but differ in \emph{what} they supervise. Several optimize the reported map itself---CAM-loss \citep{wang2021camloss}, contrastive Grad-CAM consistency \citep{pillai2022cgc}, and, closest to our medical extension, explanation-supervised biomedical training \citep{faruqui2026eal}. Others instead constrain an attribution-\emph{derived} signal rather than the map they report: Grad-CAM-guided channel--spatial attention \citep{xu2021gradcamcsa} uses Grad-CAM-derived information to guide an attention module, and input-gradient and explanation regularization \citep{ross2017right,schramowski2020rrsr} penalize gradients or explanations outside the annotated region. Our CAM-Align, CAM-Dice, and CAM-BCE controls are the matched direct-supervision baselines. \emph{Attention and routed representations} instead expose the supervised spatial object on the prediction path: attention-branch and guided-attention methods \citep{fukui2019abn,li2018tell}, a spatially attentive output layer \citep{kim2020saol}, and bounded logit attention \citep{baumhauer2021bla} all make the reported map part of the classifier's own computation. BGER differs from both: it supervises a class-agnostic route through which classification flows and evaluates a separately computed pre-gate predicted-class Grad-CAM that never enters the objective. No individual component is new in isolation; the contribution is this combination, together with the matched no-routing and detach interventions that measure what routing contributes and whether the gain depends on gradients through the gate. Our equal-annotation ladder compares these settings directly: the same box supplied as input masking, an attention branch, or a route produces very different Grad-CAM maps (App.~Tab.~\ref{tab:annotation-ladder}). Calibration assesses confidence--correctness agreement \citep{guo2017calibration,naeini2015ece} but does not measure spatial overlap.

\paragraph{CAM and localization evaluation.}
Class activation maps and Grad-CAM \citep{zhou2016cam,selvaraju2017gradcam} visualize class-discriminative regions after training; Grad-CAM++ \citep{chattopadhyay2018gradcampp}, HiResCAM \citep{draelos2020hirescam}, and Score-CAM \citep{wang2020scorecam} use different weighting rules. WSOL evaluates localization from image-level supervision \citep{choe2020wsol}, whereas BGER uses training-time boxes or masks and is therefore not image-label-only WSOL. Our primary evaluation uses predicted-class post-hoc Grad-CAM; true-class dedicated WSOL maps are reported only as separately labeled localization references.

\section{Method}

\paragraph{Evidence-routed classification.}
Each sample carries an optional evidence annotation $B$ (a box or mask) marking the spatially appropriate region. BGER routes classification features through a gate supervised to concentrate inside $B$, while the audited Grad-CAM stays out of the loss: the annotation supervises \emph{where} features are routed, not \emph{what} post-hoc heatmap should be drawn. At inference no annotation is used and BGER deploys exactly like the CE classifier.

Let $\mathcal{D}=\{(x_i,y_i,a_i,B_i)\}_{i=1}^{N}$, where $a_i\in\{0,1\}$ marks a valid annotation and by convention $B_i\equiv0$ when $a_i=0$, and let $U=hw$. With backbone $F_\theta$, class-agnostic $1{\times}1$ gate $v\in\mathbb{R}^{C}$, $b\in\mathbb{R}$, and linear head $W\in\mathbb{R}^{K\times C}$, $c\in\mathbb{R}^{K}$,
\begin{equation}
\label{eq:forward}
\begin{aligned}
H_i &= F_\theta(x_i)\in\mathbb{R}^{C\times h\times w}, &
g_{i,u} &= v^{\top}H_{i,u}+b,\\
M_{i,u} &= \sigma(g_{i,u}), &
z^{\mathrm{route}}_i &= \tfrac{1}{U}\textstyle\sum_{u} M_{i,u}H_{i,u},
\end{aligned}
\end{equation}
and $s_i=Wz^{\mathrm{route}}_i+c$: a pre-softmax logit vector over $K$ classes for multiclass data, and a scalar positive-class logit ($K{=}1$) for the binary chest-X-ray protocols.

\paragraph{Balanced-BCE route.}
BCE-BGER binarizes the annotation and nearest-neighbor downsamples it to the gate grid, $B_i\in\{0,1\}^{1\times h\times w}$. With $p_i=\operatorname{clip}(U^{-1}\sum_u B_{i,u},p_{\min},1-p_{\min})$ and $\bar{M}_{i,u}=\operatorname{clip}(M_{i,u},\epsilon,1-\epsilon)$,
\begin{equation}
\label{eq:bbce}
\begin{split}
\ell^{(i)}_{\mathrm{BBCE}}=-\frac{1}{2U}\sum_{u}\Big[&\frac{B_{i,u}}{p_i}\log\bar{M}_{i,u}\\[-2pt]
&+\frac{1-B_{i,u}}{1-p_i}\log\big(1-\bar{M}_{i,u}\big)\Big],
\end{split}
\end{equation}
\begin{equation}
\label{eq:obj}
\mathcal{L}_{\mathrm{BGER}}=\frac{1}{N}\sum_{i}\Big[\ell_{\mathrm{cls}}(s_i,y_i)+a_i\,\lambda_{\mathrm{route}}\,\ell^{(i)}_{\mathrm{BBCE}}\Big].
\end{equation}
When clipping is inactive the two terms assign equal aggregate weight to foreground and background pixels; clipping caps extreme weights for very small or near-global evidence regions ($p_{\min}=0.02$, $\epsilon=10^{-6}$). Here $\ell_{\mathrm{cls}}$ is multiclass cross-entropy for natural images and weighted binary cross-entropy for the chest-X-ray protocols, and $a_i=0$ leaves $\ell_{\mathrm{cls}}$ only. Dice-BGER and the separately weighted Energy-BGER objective are defined in App.~\ref{sec:training-recipes}.

\paragraph{Matched mechanism interventions.}
Setting $z^{(q)}_i=U^{-1}\sum_u\phi_q(M_{i,u})H_{i,u}$ with $\phi_{\mathrm{route}}(m)=m$, $\phi_{\mathrm{aux}}(m)=1$ and $\phi_{\mathrm{res}}(m)=1+\eta m$ ($\eta=0.5$), each arm trains
\begin{equation}
\label{eq:paths}
\mathcal{L}^{(q)}=\tfrac{1}{N}\textstyle\sum_i\big[\ell_{\mathrm{cls}}(W_qz^{(q)}_i+c_q,y_i)+a_i\lambda_{\mathrm{route}}\ell^{(i)}_{\mathrm{BBCE}}\big].
\end{equation}
All three share the gate parameterization, evidence target, route objective and $\lambda_{\mathrm{route}}$, differing only in how the gate enters the classification path; the auxiliary arm still backpropagates the route loss into the shared backbone through $M$. Arms are trained independently (arm-specific parameters suppressed), so the comparison localizes where the effect enters rather than apportioning it additively.

\paragraph{Audited object and detach test.}
The audited object is the post-hoc Grad-CAM read at the \emph{pre-gate} features $H$ for the predicted class $\hat{k}=\arg\max_j s_j$ (decision-conditioned, not a GT-known map); the chest-X-ray protocols instead fix $k$ to the positive-abnormality logit. The audit reads $H$ upstream of the gate, but the class score still reaches $H$ through it. Suppressing the sample index, and for the GAP--linear instantiation,
\begin{equation}
\label{eq:cam}
\begin{split}
A_k(H)&=\mathrm{ReLU}\Big(\textstyle\sum_{c}\alpha_{k,c}H_c\Big),\;\;
\alpha_{k,c}=\tfrac{1}{U}\textstyle\sum_{u}\tfrac{\partial s_k}{\partial H_{c,u}},\\[-1pt]
\tfrac{\partial s_k}{\partial H_{c,u}}&=\tfrac{1}{U}\big[M_u w_{k,c}+\sigma'(g_u)(w_k^{\top}H_u)v_c\big],\\[-1pt]
\boldsymbol{\alpha}_k&=\mu\,w_k+\beta_k\,v,\quad \mu=U^{-2}\textstyle\sum_u M_u>0,
\end{split}
\end{equation}
$\beta_k=U^{-2}\sum_u\sigma'(g_u)(w_k^{\top}H_u)$ (derivation in the appendix). The detach test re-reads the same trained model with $s^{\mathrm{det}}_k=U^{-1}\sum_u w_k^{\top}[\mathrm{sg}(M_u)H_u]+c_k$, where $\mathrm{sg}$ keeps the forward value and has zero derivative; hence $s^{\mathrm{det}}_k=s_k$ in the forward pass---only the audit's gradient path changes---while $\boldsymbol{\alpha}^{\mathrm{det}}_k=\mu\,w_k$, and under the positive-scale-invariant normalization used by the evaluator $\mu$ cancels, leaving the normalized head-weight CAM. This is asserted only for the GAP--linear case; Swin-T \citep{liu2021swin} and ConvNeXt-T \citep{liu2022convnext} are checked empirically.

Medical spatial metrics are therefore finding-conditioned lesion grounding on positive boxed studies, not predicted-class grounding; external WSOL references, marked separately, provide true-class localization maps. The second spatial object is the learned gate $M$, a diagnostic routing map reported separately and never assigned classifier accuracy or calibration.

\section{Experiments}

\paragraph{Datasets.}
We use CUB-200-2011 \citep{wah2011cub}, Oxford-IIIT Pet \citep{parkhi2012pets}, and Stanford Dogs \citep{khosla2011dogs} as main natural-image evidence. CUB and Dogs provide object boxes; Pet provides foreground trimaps, from which we use the full foreground mask as the primary evidence target and derive boxes only for box metrics. We also include two chest X-ray medical extensions with radiologist-provided boxes: VinDr-CXR \citep{nguyen2022vindr} (multi-abnormality boxes) and RSNA Pneumonia 2018 \citep{rsna2018pneumonia} (lung-opacity boxes). COCO-SO, a controlled single-target subset derived from COCO \citep{lin2014coco}, is appendix-only because it is seed-0 and mixed.

\paragraph{Methods.}
We compare CE; BCE-BGER; the Dice-BGER and Energy-BGER route variants; and three direct-proxy controls---CAM-Align, CAM-Dice, and CAM-BCE---which apply an energy, soft-Dice, or balanced-BCE objective directly to a CAM proxy rather than to a route (App.~\ref{app:proxy}). Equal-annotation controls include input masking, an ABN-style supervised-attention branch \citep{fukui2019abn}, residual routing, and frozen-backbone routing. The learned gate is reported only as a diagnostic map.

\paragraph{Metrics.}
Predictive metrics are accuracy and ECE (AUROC/AUPRC for chest X-ray); spatial metrics are Energy-in-Box/Mask (EIB/EIM), Pointing Game (PG; \citealp{zhang2016excitation}), and MaxBoxAccV2. The primary CUB table, the BCE-BGER mechanism controls, and the high-confidence spatially-ungrounded rate (HCU) all use the full $5{,}794$-image test set. Region-restricted corruption uses a fixed $N=1{,}000$ subset; operator consistency uses a separately fixed $N=500$ subset. Medical spatial metrics use positive boxed test images only. Code and per-sample records are provided as a code-and-data supplement accompanying this paper.

\paragraph{Implementation details.}
For CUB, Pet, and Dogs we use ImageNet-initialized ResNet-50 \citep{he2016resnet}; the chest X-ray protocols use DenseNet-121 \citep{huang2017densenet} with weighted BCE; seeds 0--2 when marked three-seed (full recipes: App.\ Sec.~\ref{sec:training-recipes}). Hyperparameters are selected on seed-0 \emph{validation}, never on test. BCE-BGER uses $\lambda_{\mathrm{route}}=2$ on CUB and Dogs; soft Dice uses the same weight. Energy--area results retain their separately selected recipe and are labeled as an objective variant. CAM-BCE applies the objective-matched balanced BCE to a sigmoid CAM proxy on the same backbone. Internal classifier rows report predicted-class Grad-CAM. BGER adds $C{+}1=2{,}049$ parameters ($+0.009\%$) and one $1{\times}1$ convolution at inference.

\subsection{BCE-BGER Route-to-CAM Transfer on CUB and Dogs}

Across three seeds, BCE-BGER improves full-test CUB Grad-CAM MaxBoxAccV2 from $0.584$ to $0.715$ (Tab.~\ref{tab:bger-cub-external}); accuracy/ECE are $.748/.016$ versus CE $.749/.017$. A paired image-level bootstrap over all $5{,}794$ test images ($10{,}000$ resamples) puts the MaxBoxAccV2 gain over CE at $+.130$ ($95\%$ CI $[.126,.134]$, excludes zero) with accuracy indistinguishable from CE ($\Delta$ CI $[-.007,.005]$); BCE-BGER also exceeds the Energy ($+.057$ $[.053,.059]$) and Dice ($+.024$ $[.022,.027]$) route variants, with seed-level complete separation (App.~Tab.~\ref{tab:perseed-separation}).

The result replicates on Stanford Dogs (App.~Tab.~\ref{tab:dogs-full}): BCE-BGER raises MaxBoxAccV2 from CE $.757$ to $.832$ (Energy-BGER $.801$; Dice-BGER $.829$) at similar accuracy ($.878$ vs.\ $.880$) and calibration, and all six BCE- and Dice-BGER seeds exceed all three Energy-BGER seeds.

\begin{table*}[t]
\centering
\small
\setlength{\tabcolsep}{3.6pt}
\begin{tabular}{lllrrrrr}
\toprule
Family & Method & Supervised object & Acc & ECE15 & EIB & PG & MaxBoxAccV2 \\
\midrule
Label-only & CE & --- & $.749{\pm}.005$ & $.017{\pm}.004$ & $.788$ & $.966$ & $.584{\pm}.001$ \\
\midrule
Direct attribution & CAM-Align & CAM proxy (energy) & $.747{\pm}.003$ & $.022{\pm}.003$ & $\mathbf{.866}$ & $.996$ & $.651{\pm}.005$ \\
 & CAM-Dice & CAM proxy ($\sigma(S)$, Dice) & $.748{\pm}.003$ & $.016{\pm}.007$ & $.786$ & $.984$ & $.696{\pm}.006$ \\
 & CAM-BCE & CAM proxy ($\sigma(S)$, BCE) & $.737{\pm}.005$ & $.030{\pm}.005$ & $.837$ & $\mathbf{.997}$ & $\mathbf{.769}{\pm}.008$ \\
\midrule
Feature routing & Energy-BGER & internal route (energy--area) & $.750{\pm}.001$ & $\underline{.015}{\pm}.002$ & $.859$ & $.992$ & $.658{\pm}.001$ \\
 & Dice-BGER & internal route (soft Dice) & $.749{\pm}.001$ & $\mathbf{.013}{\pm}.002$ & $.829$ & $.991$ & $.690{\pm}.005$ \\
 & Residual-BCE & internal route (bypass, BCE) & $\mathbf{.755}{\pm}.003$ & $.019{\pm}.007$ & $.814$ & $.987$ & $.714{\pm}.003$ \\
 & \textbf{BCE-BGER} & internal route (balanced BCE) & $.748{\pm}.001$ & $.016{\pm}.004$ & $.840$ & $.995$ & $\underline{.715}{\pm}.005$ \\
\bottomrule
\end{tabular}
\caption{Objective-matched CUB comparison on the full test set ($N{=}5{,}794$; three seeds; predicted-class Grad-CAM), grouped by supervision family. Direct-proxy methods supervise an attribution-family map; route methods supervise an internal feature route and never place an attribution map in the loss. BCE-BGER uses class-balanced binary cross-entropy for route supervision; CAM-BCE applies the same loss to a CAM proxy and reaches higher localization at lower accuracy (full frontier: Fig.~\ref{fig:pareto}).}
\label{tab:bger-cub-external}
\end{table*}

\subsection{Route Supervision versus Direct CAM Supervision}

CAM-BCE directly optimizes a CAM proxy and reaches $0.769$ on CUB, higher on localization ($\Delta$ over BCE-BGER $+.054$, CI $[.051,.058]$) but at a real accuracy cost (BCE-BGER accuracy is higher, $\Delta$ $+.011$, CI $[.005,.016]$). BCE-BGER instead keeps the supervised route distinct from the map it is evaluated on.

\begin{figure}[tb]
    \centering
    \includegraphics[width=\columnwidth]{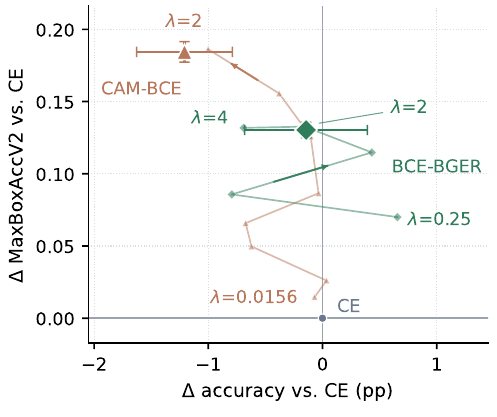}
    \caption{\textbf{CE-relative accuracy--localization trade-offs on CUB.} Axes show accuracy change from CE in percentage points and MaxBoxAccV2 change from CE. Small markers are observed seed-0 weight-sweep outcomes connected in increasing-$\lambda$ order; large markers with error bars are the validation-selected $\lambda{=}2$ means and standard deviations of paired CE-relative differences across three matched seeds. Gate-route BCE-BGER yields $+.130$ MaxBoxAccV2 at approximately $-0.1$ pp accuracy, whereas direct-proxy CAM-BCE yields $+.184$ at approximately $-1.2$ pp. Lines connect discrete trained models only; neither family dominates over the tested grids.}
    \label{fig:pareto}
\end{figure}

\paragraph{Accuracy--localization frontier.}
Both families are swept under the same validation-only protocol (matched accuracy constraint, evaluator, and box target; test used once per point; full per-point table in App.~\ref{app:frontier}), and the two observed sweeps are mutually non-dominating (Fig.~\ref{fig:pareto}; swept points are seed-0, the validation-selected points are confirmed with three seeds). The gate's frontier rises with route weight ($.654\!\to\!.717$ for $\lambda\in\{0.25,\ldots,2\}$) and then saturates ($\lambda{=}4$ gives $.716$) while holding accuracy; the proxy passes through a comparable point at the same accuracy and extends beyond it to $.769$ at accuracy $.739$. Direct proxy supervision maximizes localization, while the route preserves more accuracy.

\subsection{Spatial Supervision Drives Most Transfer; Routing Adds a Smaller Increment}

All controls here use BCE-BGER, three seeds, and the full-test evaluator of Tab.~\ref{tab:bger-cub-external} (Tab.~\ref{tab:bce-controls}). \emph{Frozen backbone}: retraining only the gate and head under the BCE route loss reaches MaxBoxAccV2 $.597{\pm}.000$---about a tenth of the full $.584\!\to\!.715$ gain---so the transfer requires the backbone representation to reshape. \emph{Shuffled boxes}: replacing each box with another image's box retains $.619{\pm}.003$, about a quarter of the gain, so correct object extents matter rather than a generic center prior.

\emph{Auxiliary BCE, no routing}: the no-routing control supervises the identical gate with the identical balanced-BCE loss at the same $\lambda_{\mathrm{route}}$, but classifies from the ungated $\mathrm{GAP}(H)$, so the backbone still receives the spatial gradient while the supervised map never enters the classification path. This reaches $.683{\pm}.001$ ($+.099$ over CE) at accuracy indistinguishable from BCE-BGER ($-.003{\pm}.003$). Spatial supervision applied through the gate therefore transfers primarily by reshaping the shared backbone \emph{even when classification bypasses the gate}, and routing is not necessary for most of the effect. Routing adds a smaller $+.032$ increment, consistent across the three seeds and at no accuracy cost; the route need not be exclusive, however, since a non-exclusive residual route reaches $.714{\pm}.003$, essentially matching the exclusive $.715{\pm}.005$ (per-seed increments, intervals, and the Energy contrast in App.~\ref{app:bce-controls}). The BCE results thus support an additional routing effect but not a benefit from exclusive gating. ``Ret.'' is descriptive: these interventions train different systems and need not compose additively.

For the GAP--linear ResNet instantiation, the Grad-CAM weights decompose exactly as $\boldsymbol{\alpha}_k=\mu w_k+\beta_k v$ with $v$ an image-independent gate direction (App.~Eq.~\ref{eq:alpha-decomp}; measured projection residual $1.4\times10^{-7}$). We test the consequence directly. Detaching the gate derivative---severing $\partial M/\partial H$ while preserving the forward gate values---changes MaxBoxAccV2 by only $.0029$ ($.715\!\to\!.712$), retains $97.7\%$ of the CE$\to$BCE-BGER gain, and yields a mean per-image map correlation of $.9999$, with $99.99\%$ of images above $.98$. The detached map is numerically identical to the head-weight CAM, confirming the decomposition's prediction that the remaining image-dependent scalar cancels under normalized visualization. Gradients through the gate therefore contribute little to this map for the GAP--linear ResNet instantiation: the gain is carried by the reshaped features, not by the gate imprinting a template on the map.

\begin{table*}[tb]
\centering
\small
\setlength{\tabcolsep}{4pt}
\begin{tabular}{llrrrrr}
\toprule
Variant & Class.\ path & Acc & ECE15 & MaxBoxV2 & $\Delta$ CE & Ret. \\
\midrule
CE & $H$ & $.749{\pm}.005$ & $.017{\pm}.004$ & $.584{\pm}.001$ & --- & $0\%$ \\
Frozen backbone & $M\odot H$ & $.731{\pm}.006$ & $.026{\pm}.017$ & $.597{\pm}.000$ & $+.013$ & $10\%$ \\
Shuffled boxes & $M\odot H$ & $.734{\pm}.007$ & $.018{\pm}.008$ & $.619{\pm}.003$ & $+.035$ & $27\%$ \\
Aux.\ BCE, no routing & $H$ & $.751{\pm}.003$ & $.012{\pm}.003$ & $.683{\pm}.001$ & $+.099$ & $76\%$ \\
Residual-BCE & $H\odot(1{+}\eta M)$ & $.755{\pm}.003$ & $.019{\pm}.007$ & $.714{\pm}.003$ & $+.130$ & $99\%$ \\
\textbf{BCE-BGER} & $M\odot H$ & $.748{\pm}.001$ & $.016{\pm}.004$ & $.715{\pm}.005$ & $+.131$ & $100\%$ \\
\bottomrule
\end{tabular}
\caption{BCE-BGER mechanism controls (CUB full test, three seeds, same evaluator as Tab.~\ref{tab:bger-cub-external}). ``Gain ret.'' is the fraction of the CE$\to$BCE-BGER MaxBoxAccV2 gain retained; it is descriptive, and the interventions do not compose additively. \emph{Aux.\ BCE, no routing} applies the identical gate and box loss but classifies from the ungated $\mathrm{GAP}(H)$; it isolates shared-backbone reshaping from routing and recovers most of the gain. ``Class.\ path'' is the representation the classifier reads. Full protocol and equal-annotation controls: App.~\ref{app:bce-controls},~\ref{app:energy-diagnostics}.}
\label{tab:bce-controls}
\end{table*}

Two further controls bound the effect. Both are \emph{legacy Energy-BGER diagnostics} run under the earlier Energy recipe and its evaluator (whose CE reference reads $.588$ rather than $.584$); they are not BCE-BGER ablations and should not be read as such. Under an equal-annotation protocol, supplying the identical box through input masking, randomization, or cropping leaves the resulting map near CE, and an ABN-style attention branch moves it only slightly: feature routing produces the largest shift among the tested equal-annotation non-proxy controls (App.~Tab.~\ref{tab:annotation-ladder}). Removing the box loss collapses localization to CE level, and an Energy-BGER annotation-budget diagnostic on that fixed protocol recovers over half the gain with $10\%$ of boxes (App.~Tab.~\ref{tab:fraction-curve}).

\subsection{Regional Sensitivity under Corruption}

\begin{figure}[tb]
    \centering
    \includegraphics[width=\columnwidth]{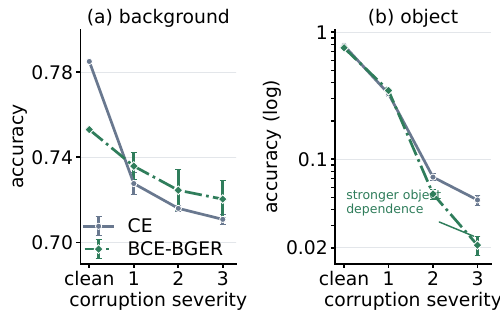}
    \caption{Region-restricted corruption for BCE-BGER (CUB, fixed $N{=}1{,}000$ subset, three seeds, matched noise+blur). BCE-BGER retains more accuracy than CE when only the background is corrupted (a) and loses more when only the annotated object is corrupted (b), consistent with greater object-region sensitivity. Corruption probes sensitivity, not causal reliance.}
    \label{fig:dissociation}
\end{figure}

Corrupting only the annotated object or its complement tests regional sensitivity without reading a CAM (Fig.~\ref{fig:dissociation}). At severity 3, BCE-BGER is less affected than CE by background corruption ($.720$ vs.\ $.711$) and more affected by object corruption ($.021$ vs.\ $.048$). On the full test set, high-confidence spatially-ungrounded predictions (confidence $\ge.9$ with EIB $<.5$) fall twenty-fold, from $1.24{\pm}.02\%$ for CE to $0.06{\pm}.01\%$ for BCE-BGER.

\subsection{Lesion Grounding on Chest X-rays}

Both chest-X-ray protocols use DenseNet-121 with radiologist boxes as evidence targets; spatial metrics use positive boxed images only (Tab.~\ref{tab:ext-datasets}). Lesions are often off-center, irregular, and different in scale and geometry from the object-centric natural-image targets, so this is a different evaluation regime. For these DenseNet-121 checkpoints, detaching the gate-gradient path retains $94\%$ of the RSNA gain and $78\%$ of the VinDr gain (App.~Tab.~\ref{tab:detach-densenet}), excluding the Swin-like collapse discussed below.

On \textbf{VinDr-CXR}, BCE-BGER improves its Grad-CAM on both spatial axes over CE ($.142\!\to\!.209$, EIB $.147\!\to\!.263$), but direct CAM supervision is clearly stronger spatially ($.361$/$.402$) and has the higher AUROC point estimate ($.885$ vs.\ $.862$, unresolved at three seeds): VinDr favors direct CAM supervision. On \textbf{RSNA} the ordering partly reverses---BCE-BGER gives the highest MaxBoxAccV2 ($.427$) and AUROC ($.988$ vs.\ $.986$, also unresolved), whereas CAM-BCE gives stronger EIB/PG ($.568$/$.782$ vs.\ $.379$/$.713$). Which method looks better depends on which spatial metric is read---the same accuracy--localization tension as the CUB frontier, now \emph{between} metrics rather than along a sweep.

\begin{table*}[tb]
\centering
\small
\setlength{\tabcolsep}{6pt}
\begin{tabular}{llrrrrr}
\toprule
Data & Method & AUROC & AUPRC & ECE15 & EIB & MaxBoxV2 \\
\midrule
VinDr & CE & $.851{\pm}.021$ & $.768{\pm}.027$ & $.138{\pm}.012$ & $.147{\pm}.012$ & $.142{\pm}.015$ \\
 & \textit{CAM-Align} & $.878{\pm}.027$ & $.805{\pm}.028$ & $\mathbf{.085}{\pm}.013$ & $.390{\pm}.006$ & $.255{\pm}.044$ \\
 & BCE-BGER & $.862{\pm}.016$ & $.805{\pm}.023$ & $.125{\pm}.020$ & $.263{\pm}.029$ & $.209{\pm}.011$ \\
 & \textit{CAM-BCE} & $\mathbf{.885}{\pm}.011$ & $\mathbf{.820}{\pm}.013$ & $.103{\pm}.031$ & $\mathbf{.402}{\pm}.015$ & $\mathbf{.361}{\pm}.002$ \\
\midrule
RSNA & CE & $.985{\pm}.002$ & $.983{\pm}.003$ & $.027{\pm}.007$ & $.189{\pm}.036$ & $.200{\pm}.066$ \\
 & \textit{CAM-Align} & $.986{\pm}.001$ & $.983{\pm}.001$ & $.024{\pm}.011$ & $.565{\pm}.017$ & $.361{\pm}.018$ \\
 & BCE-BGER & $\mathbf{.988}{\pm}.001$ & $\mathbf{.986}{\pm}.001$ & $.014{\pm}.012$ & $.379{\pm}.013$ & $\mathbf{.427}{\pm}.029$ \\
 & \textit{CAM-BCE} & $.986{\pm}.002$ & $.983{\pm}.004$ & $\mathbf{.011}{\pm}.005$ & $\mathbf{.568}{\pm}.011$ & $.420{\pm}.015$ \\
\bottomrule
\end{tabular}
\caption{Controlled chest-X-ray lesion grounding (three seeds, mean$\pm$std; DenseNet-121). Both protocols use a single abnormality logit; the CAM is conditioned on that logit and spatial metrics are computed on positive boxed test images only, so---unlike the predicted-class CUB evaluation---these rows read the map for a known-present finding. Discriminative performance is reported as AUROC/AUPRC because both protocols are class-imbalanced (accuracy, PG, and AdaECE in App.~Tab.~\ref{tab:medical-full}). Italics denote direct CAM-proxy supervision; BCE-BGER and CAM-BCE use the objective-matched balanced BCE ($\lambda_{\mathrm{route}}{=}2$).}
\label{tab:ext-datasets}
\end{table*}

\paragraph{When soft energy and box overlap diverge.}
A per-case subgroup analysis (App.~Tab.~\ref{tab:rsna-subgroup-bce}) shows that the route-vs-proxy ordering is a metric property rather than a geometry effect: CAM-BCE's soft-energy advantage over BCE-BGER is nearly constant across lesion size and box count, while their thresholded box overlap is indistinguishable everywhere. Direct proxy supervision buys concentration on the map it optimizes, uniformly, without a matching overlap gain, so the two families can swap ranks between soft energy-in-box and thresholded MaxBoxAccV2 while staying close on the latter. Lesion grounding thus supports transfer beyond object-centric images and shows that soft concentration and thresholded localization are not interchangeable.

\subsection{CAM Readouts and Backbone Dependence}

\begin{table}[t]
\centering
\small
\setlength{\tabcolsep}{6pt}
\begin{tabular}{lrrr}
\toprule
Readout & CE & BCE-BGER & \textit{CAM-BCE} \\
\midrule
Grad-CAM & $.593$ & $.728$ & \textit{.778} \\
Grad-CAM++ & $.602$ & $.736$ & \textit{.804} \\
HiResCAM & $.593$ & $.728$ & \textit{.778} \\
Score-CAM-128 & $.590$ & $.734$ & \textit{.798} \\
\bottomrule
\end{tabular}
\caption{Localization (MaxBoxAccV2) across CAM readouts on the fixed $N{=}500$ CUB subset (three seeds, identical images and checkpoints). BCE-BGER improves over CE under all four readouts, including gradient-free Score-CAM-128; CAM-BCE (italic, direct supervision) remains higher, consistent with Fig.~\ref{fig:pareto}. This establishes consistency across readout rules, not attribution-method independence, because all operators read the same box-supervised representation. Full EIB/PG in App.~Tab.~\ref{tab:bce_operator_consistency}.}
\label{tab:readouts}
\end{table}

The localization shift is not specific to the Grad-CAM weighting rule. On the fixed $N{=}500$ CUB subset, BCE-BGER exceeds CE under Grad-CAM, Grad-CAM++, HiResCAM, and gradient-free Score-CAM-128 (Tab.~\ref{tab:readouts}), with MaxBoxAccV2 gains of about $.13$--$.14$; CAM-BCE remains higher under every readout, consistent with the trade-off in Fig.~\ref{fig:pareto}. Because all operators read the same box-supervised representation, this establishes consistency across readout rules, not attribution-method independence.

The detach experiment addresses a different question: whether the standard Grad-CAM gain itself depends on gradients through the gate (Tab.~\ref{tab:arch-probe}). On ResNet-50 the gain survives detachment and comes mostly from representation reshaping: the no-routing arm reaches $.683$, $76\%$ of the CE-to-BCE-BGER gain, and detaching the gate gradient changes the routed map only marginally ($.715$ to $.712$, $r{=}.9999$). The two DenseNet-121 medical settings behave similarly (Tab.~\ref{tab:arch-probe}b): they retain $94\%$ and $78\%$ of their gains at map correlations $.957$ and $.940$, so the reported medical localization gain also survives removal of the gate-gradient derivative---though, unlike the ResNet-50 no-routing arm, this does not separate reshaping from routing.

The other two backbones behave differently. On Swin-T the standard routed Grad-CAM does not survive detachment. It reaches $.896$ versus $.280$ for CE, but detaching $\partial M/\partial H$ drops it to $.256$---below CE---and the standard and detached maps are anti-correlated ($r{=}{-}.16$), with the standard map $.91$-correlated with the gate itself. Its apparent gain therefore came from gradients through the gate (gate-gradient imprinting), not from a reshaped representation; the no-routing figure is only $+.038$. Prediction is unaffected, and detachment removes only the standard map's standing as evidence of representation transfer. On ConvNeXt-T detachment leaves MaxBoxAccV2 essentially unchanged ($.253$ to $.255$, $r{=}.974$), but BCE-BGER does not improve localization at all: both values stay below CE ($.291$) and calibration worsens. Here the stable detach result rules out contamination from the gate gradient, so the negative result is real rather than a measurement artifact. The detach test thus passes for the reported ResNet-50 and DenseNet-121 GAP--linear implementations and should be re-checked on a new backbone. Complete BCE-BGER mechanism arms and detach results are in App.~Tab.~\ref{tab:detach-arch},~\ref{tab:detach-densenet}; App.~Tab.~\ref{tab:swin-generality} and Fig.~\ref{fig:dose-response} report separate Energy-BGER architecture sweeps.

\begin{table}[t]
\centering
\small
\setlength{\tabcolsep}{4.5pt}
\begin{tabular}{lrrrrr}
\multicolumn{6}{c}{\emph{(a) CUB backbone mechanism arms}}\\
\toprule
Backbone & CE & Aux & BGER & det. & corr \\
\midrule
ResNet-50 & $.584$ & $.683$ & $\mathbf{.715}$ & $.712$ & $.9999$ \\
Swin-T & $.280$ & $.317$ & $.896$ & $.256$ & $-.16$ \\
ConvNeXt-T & $.291$ & $.288$ & $.253$ & $.255$ & $.974$ \\
\bottomrule
\end{tabular}
\\[3pt]
\begin{tabular}{lrrrr}
\multicolumn{5}{c}{\emph{(b) DenseNet-121 medical detach stability}}\\
\toprule
Dataset & std. & det. & corr. & retained \\
\midrule
RSNA & $.427$ & $.413$ & $.957$ & $94\%$ \\
VinDr & $.209$ & $.194$ & $.940$ & $78\%$ \\
\bottomrule
\end{tabular}
\caption{\textbf{(a)}~Predicted-class Grad-CAM MaxBoxAccV2 on the full CUB test set (three seeds): CE, auxiliary BCE without routing (Aux), BCE-BGER, BCE-BGER with the gate gradient detached (det.), and the mean per-image standard-vs-detached map correlation (corr). \textbf{(b)}~Positive-finding Grad-CAM MaxBoxAccV2 on positive boxed chest-X-ray studies (three seeds): standard vs.\ detached MaxBoxAccV2, map correlation, and the fraction of the CE$\to$BCE-BGER gain retained. Read values within each panel, not across them. Panel~(b) tests only whether the reported localization gain survives removing the gate-gradient derivative; it does not isolate representation reshaping from routing as the matched ResNet-50 no-routing arm does. On ConvNeXt-T both localization scores stay below CE while the standard and detached maps remain highly correlated, so the negative result is not a detachment artifact.}
\label{tab:arch-probe}
\end{table}

\paragraph{Scene and target-geometry boundaries.}
On Pet, whose loose full-object mask favors diffuse maps, the Energy route improves EIM/PG but not box IoU (App.~Tab.~\ref{tab:multi}); on cluttered COCO-SO scenes, soft evidence improves without thresholded localization (App.~Tab.~\ref{tab:coco}). Both are cases where soft evidence concentration and thresholded box localization come apart.

\section{Limitations}

BGER requires box or mask supervision and is not image-label-only WSOL. The gate $M$ is diagnostic, and improved post-hoc CAM concentration does not establish causal reliance: separating supervision from evaluation removes direct proxy fitting but not the influence of supervision on the representation, and a split-disjoint spurious-context counterexample (App.~\ref{app:spurious}) shows that CAM concentration does not imply background-shortcut robustness. The medical protocols are controlled lesion-grounding benchmarks, not clinical validation. Detachment also behaves differently across backbones: Grad-CAM is stable on ResNet-50 and the two DenseNet-121 settings, but on Swin-T the standard-map gain depends on gradients through the gate, and on ConvNeXt-T routing does not improve localization even though detachment is stable; the detach test should be repeated before interpreting results on a new backbone. Other limits---clutter, architecture, and target geometry---are reported with their experiments in the Results.

\section{Conclusion}

Balanced-BCE routing (BCE-BGER) shows that spatial supervision can improve a classifier's own separately evaluated post-hoc Grad-CAM---on CUB, Stanford Dogs, and controlled chest-X-ray lesion grounding, at similar accuracy---without optimizing the CAM itself. Matched controls indicate that most of the ResNet-50 improvement comes from reshaping the shared representation, while routing contributes a smaller additional effect. The detach results show why this distinction matters: the gain survives on ResNet-50 and the reported DenseNet-121 settings but disappears on Swin-T, so route-to-CAM transfer cannot be assumed to have the same mechanism across architectures.

\FloatBarrier
\bibliography{references}

\clearpage

\renewcommand{\topfraction}{0.92}
\renewcommand{\bottomfraction}{0.85}
\renewcommand{\textfraction}{0.07}
\renewcommand{\floatpagefraction}{0.88}
\renewcommand{\dbltopfraction}{0.92}
\renewcommand{\dblfloatpagefraction}{0.88}
\setcounter{topnumber}{3}
\setcounter{bottomnumber}{2}
\setcounter{totalnumber}{5}
\setcounter{dbltopnumber}{3}

\appendix
\setcounter{secnumdepth}{1}

\section{Novelty Matrix: Nearest Methods}
\label{app:novelty}
Table~\ref{tab:novelty} places BGER against its nearest neighbors on the axes that matter for this paper: which spatial object receives supervision, which path carries classification, which object is evaluated, whether an attribution-family quantity enters the loss, whether the spatial supervision is class-conditioned, whether supervised and evaluated objects are computationally distinct, and whether an explicit intervention tests the audit. To our knowledge, BGER differs from the closest methods by jointly combining a class-agnostic supervised route, a separately computed pre-gate predicted-class CAM, and an explicit intervention testing whether the audited gain survives removal of the gate-gradient path; no single axis is unique on its own. The direct-CAM proxies are our own objective-matched controls, showing that the same objective on a CAM-family map removes the separation while reaching higher raw localization.

\begin{table*}[t]
\centering
\small
\setlength{\tabcolsep}{3.5pt}
\begin{tabular}{llllcccc}
\toprule
Method & Supervised obj. & Class.\ path & Evaluated obj. & \shortstack{Attr.\\in loss} & \shortstack{Class\\cond.} & \shortstack{Audit\\sep.} & \shortstack{Detach\\audit} \\
\midrule
CAM/Grad-CAM & none & GAP & the CAM & --- & --- & --- & --- \\
WSOL & image label & GAP & true-cls map & no & yes & no & no \\
CAM-loss & feat.\ vs.\ CAM & GAP & CAM & yes & yes & no & no \\
CGC & CAM consist. & GAP & Grad-CAM & yes & yes & no & no \\
SAOL & output attn. & attn.\ logits & attention & no & no & no & no \\
ABN & attn.\ branch & attn.\ feat. & attention & yes & yes & no & no \\
Guided attn. & attn.\ map & attn.\ wt. & attention & yes & yes & no & no \\
RRR & input grad. & GAP & input-grad & yes & yes & no & no \\
Expl.-aware (RRSR) & explanation feedback & GAP & saliency/expl. & yes & yes & no & no \\
BLA & learned expl.\ mask & masked feat. & that mask & no & no & no & no \\
Grad-CAM CSA & Grad-CAM $\to$ attn. & attn.\ feat. & Grad-CAM & yes & yes & no & no \\
Expl.-superv.\ (EAL) & expl./saliency map & GAP & that expl.\ map & yes & yes & no & no \\
Direct proxy (ours) & CAM proxy $Q$ & GAP & pred.\ CAM & yes & yes & no & no \\
\textbf{BGER (ours)} & gate $M$ & $M{\odot}H$ & pre-gate CAM & \textbf{no} & \textbf{no} & \textbf{yes} & \textbf{yes} \\
\bottomrule
\end{tabular}
\caption{Nearest methods by supervised object, classification path, and evaluated object (citations in Related Work): CAM/Grad-CAM \citep{zhou2016cam,selvaraju2017gradcam}, WSOL \citep{choe2020wsol}, CAM-loss \citep{wang2021camloss}, CGC \citep{pillai2022cgc}, SAOL \citep{kim2020saol}, ABN \citep{fukui2019abn}, guided attention \citep{li2018tell}, RRR \citep{ross2017right}, explanation-aware training \citep{schramowski2020rrsr}, bounded logit attention \citep{baumhauer2021bla}, Grad-CAM-guided channel-spatial attention \citep{xu2021gradcamcsa}, and explanation-supervised biomedical training \citep{faruqui2026eal}. ``Attr.\ in loss'': an attribution-family map (CAM, saliency, or input-gradient) appears in the objective; ``Class cond.'': the supervised spatial object is class-conditioned; ``Audit sep.'': the evaluated map is a distinct computational object from the supervised one; ``Detach audit'': an explicit intervention tests that the evaluated gain survives removing the supervised object's gradient path. Among the compared methods, BGER uniquely combines a class-agnostic supervised route, a separately computed pre-gate predicted-class CAM, and an explicit detach intervention testing audit validity.}
\label{tab:novelty}
\end{table*}

\section{Training Recipes}
\label{sec:training-recipes}
For CUB, Pet, and Dogs we use ImageNet-initialized ResNet-50, $224^2$ inputs, 15 epochs, batch 64, and SGD (momentum $0.9$, weight decay $10^{-4}$, Nesterov) with cosine decay from learning rate $0.01$. BCE-BGER and the soft-Dice variant use $\lambda_{\mathrm{route}}=2$ on CUB and Dogs. The original Energy-BGER variant uses $(\rho{=}0.25,\lambda_{\mathrm{box}}{=}1,\lambda_{\mathrm{area}}{=}0.05)$ selected on seed-0 validation; Pet and the annotation/topology diagnostics retain this Energy recipe. The chest-X-ray BCE-BGER and CAM-BCE experiments use DenseNet-121, $320^2$ inputs, AdamW, weighted classification BCE, and $\lambda_{\mathrm{route}}=2$. Energy-BGER medical analyses are labeled where retained. Test labels never enter hyperparameter selection.

\paragraph{Objective variants.}
Dice-BGER replaces the route term of Eq.~\ref{eq:obj} by $\ell^{(i)}_{\mathrm{Dice}}=1-2\sum_uM_{i,u}B_{i,u}/(\sum_uM_{i,u}+\sum_uB_{i,u}+\epsilon)$ at the same $\lambda_{\mathrm{route}}$. Energy-BGER is weighted separately rather than through $\lambda_{\mathrm{route}}$: with the inside-energy term $\ell^{(i)}_{\mathrm{in}}=1-\sum_uM_{i,u}B_{i,u}/(\sum_uM_{i,u}+\epsilon)$ it minimizes $\frac{1}{N}\sum_i\ell_{\mathrm{cls}}(s_i,y_i)+\lambda_{\mathrm{box}}\frac{1}{N}\sum_i\ell^{(i)}_{\mathrm{in}}+\lambda_{\mathrm{area}}(\langle M\rangle-\rho)^2$, where $\langle M\rangle$ is the mean gate activation over the batch and the spatial grid, as implemented. All experiments run on a single NVIDIA A100 (40\,GB) under Linux with PyTorch 2.4.1, torchvision 0.19.1 (CUDA 11.8), and Python 3.11; a $224^2$ ResNet-50 run trains in about two minutes and the $320^2$ DenseNet-121 medical runs are longer, so the full experimental suite (all datasets, backbones, three seeds, ablations, sweeps, and evaluation/bootstrap) is on the order of 20--40 GPU-hours. The reproducibility scripts that rebuild the reported numbers from the released per-seed CSVs need no GPU and run on CPU with only NumPy and pandas (pinned in the code supplement's \texttt{requirements.txt}); CPU model is therefore not load-bearing for reconstruction.

\paragraph{Transformer and ConvNeXt gate interface.}
The generic formulation (Eq.~\ref{eq:forward}) identifies the gated tensor $H$ and the pre-gate Grad-CAM readout with the same feature map, which holds exactly for the GAP--linear ResNet-50, DenseNet-121, and ConvNeXt-T backbones. For Swin-T the two are separated by the final-stage LayerNorm: Grad-CAM is hooked at the last Swin stage's \emph{pre-norm} feature tensor (reshaped to $B{\times}C{\times}h{\times}w$), whereas the routing gate consumes its \emph{post-norm} $B{\times}C{\times}h{\times}w$ representation. The audited tensor is thus still upstream of the gate, but is not literally the tensor the gate reads; this interface gap is one reason the exact GAP--linear detach decomposition is asserted only for the GAP--linear case, with Swin-T treated empirically.

\section{Proper Scoring Rules (NLL and Brier)}

Table~\ref{tab:proper-scoring} reports binning-free NLL and Brier scores. The BCE-BGER and CAM-BCE rows use the same checkpoints as the main tables; Energy-BGER and CAM-Align rows use the Energy recipe.

\begin{table}[htb]
\centering
\small
\setlength{\tabcolsep}{3pt}
\begin{tabular}{llrr}
\toprule
Dataset & Method & NLL $\downarrow$ & Brier $\downarrow$ \\
\midrule
CUB       & CE                & $.922{\pm}.009$ & $.352{\pm}.005$ \\
          & CAM-Align$^\dagger$ & $0.936$           & $0.354$           \\
          & Energy-BGER       & $\mathbf{.920}{\pm}.011$ & $\mathbf{.350}{\pm}.003$ \\
          & BCE-BGER          & $.920{\pm}.008$ & $.353{\pm}.002$ \\
          & CAM-BCE           & $.942{\pm}.007$ & $.362{\pm}.004$ \\
\midrule
Pet       & CE                & $.243{\pm}.006$ & $.109{\pm}.000$ \\
          & CAM-Align         & $.250{\pm}.010$ & $.113{\pm}.003$ \\
          & Energy-BGER       & $\mathbf{.241}{\pm}.011$ & $\mathbf{.108}{\pm}.003$ \\
\midrule
Dogs      & CE                & $\mathbf{.432}{\pm}.001$ & $\mathbf{.184}{\pm}.005$ \\
          & CAM-Align         & $.446{\pm}.010$ & $.190{\pm}.002$ \\
          & Energy-BGER       & $.447{\pm}.005$ & $.190{\pm}.002$ \\
          & BCE-BGER          & $.435{\pm}.006$ & $.186{\pm}.004$ \\
          & Dice-BGER         & $.435{\pm}.003$ & $.185{\pm}.003$ \\
\midrule
VinDr-CXR & CE                & $.738{\pm}.132$ & $.163{\pm}.003$ \\
          & CAM-Align         & $\mathbf{.480}{\pm}.020$ & $\mathbf{.134}{\pm}.008$ \\
          & Energy-BGER       & $.551{\pm}.106$ & $.137{\pm}.015$ \\
          & BCE-BGER          & $.649{\pm}.095$ & $.144{\pm}.018$ \\
          & CAM-BCE           & $.579{\pm}.080$ & $.137{\pm}.008$ \\
\midrule
RSNA      & CE                & $.176{\pm}.026$ & $.041{\pm}.001$ \\
          & CAM-Align         & $.163{\pm}.013$ & $.044{\pm}.006$ \\
          & Energy-BGER       & $.147{\pm}.014$ & $.040{\pm}.004$ \\
          & BCE-BGER          & $\mathbf{.135}{\pm}.008$ & $\mathbf{.036}{\pm}.002$ \\
          & CAM-BCE           & $.148{\pm}.014$ & $.038{\pm}.001$ \\
\bottomrule
\end{tabular}
\caption{NLL and Brier scores (lower is better; three seeds unless noted). BCE-BGER matches CE-level proper scoring on CUB and Dogs and leads on RSNA; CAM-BCE pays a small CUB cost consistent with its accuracy trade-off. $^\dagger$CUB CAM-Align is seed-0 only.}
\label{tab:proper-scoring}
\end{table}

\section{Calibration-Oriented CE Controls}
\label{app:ce-variants}

\paragraph{CE variants.}
Calibration-oriented CE variants and an evidence-preserving augmentation control leave the audited map essentially at CE level on the full test set (three seeds): focal loss \citep{lin2017focal} reaches MaxBoxAccV2 $.606{\pm}.004$ (Acc $.743$, ECE $.112$), label smoothing $.583{\pm}.006$ ($.742$/$.212$), and EPT/BMFD $.587{\pm}.003$ ($.767$/$.047$). None approaches the route or direct-proxy families, so they are reported here rather than in the main comparison.

\paragraph{Per-seed separation of the BCE-BGER comparison.}
The main-table ordering is seed-level complete (Tab.~\ref{tab:perseed-separation}): every BCE-BGER seed exceeds every Dice-BGER seed, which exceeds every Energy-BGER and CAM-Align seed; every CAM-BCE seed exceeds every BCE-BGER seed on MaxBoxAccV2 while every BCE-BGER seed exceeds every CAM-BCE seed on accuracy.

\begin{table}[htb]
\centering
\small
\setlength{\tabcolsep}{4pt}
\begin{tabular}{lccc}
\toprule
Method & \multicolumn{3}{c}{MaxBoxAccV2 (seeds 0/1/2)} \\
\midrule
CE & $.585$ & $.585$ & $.584$ \\
CAM-Align & $.653$ & $.654$ & $.646$ \\
Energy-BGER & $.656$ & $.655$ & $.657$ \\
Dice-BGER & $.686$ & $.689$ & $.696$ \\
BCE-BGER & $.717$ & $.709$ & $.718$ \\
CAM-BCE & $.771$ & $.775$ & $.760$ \\
\bottomrule
\end{tabular}
\caption{Per-seed full-test MaxBoxAccV2 for the CUB comparison. Ordering is complete across seeds: $\min$ BCE-BGER $(.709)$ $>$ $\max$ Dice-BGER $(.696)$ $>$ $\max$ Energy-BGER $(.657)$; $\min$ CAM-BCE $(.760)$ $>$ $\max$ BCE-BGER $(.718)$. On accuracy the direction reverses: $\min$ BCE-BGER $(.747)$ $>$ $\max$ CAM-BCE $(.740)$. Seed variation on the secondary metrics is likewise small: standard deviations over the same runs are EIB $.001/.001/.001/.004/.002/.011$ and PG $.002/.000/.001/.002/.001/.000$ in row order, so the EIB and PG means of Tab.~\ref{tab:bger-cub-external} are stable to the precision reported there.}
\label{tab:perseed-separation}
\end{table}

\section{BCE-BGER Mechanism Controls}
\label{app:bce-controls}

All controls use the BCE-BGER recipe (ImageNet-initialized ResNet-50, 15 epochs, $\lambda_{\mathrm{route}}{=}2$ selected on seed-0 validation), three seeds, the full $N{=}5{,}794$ test set, and the exact evaluator of the main table (predicted-class Grad-CAM at the pre-gate features, 21 thresholds, IoU $0.3/0.5/0.7$). \emph{Frozen backbone}: the backbone is initialized from the matched CE checkpoint of the same seed and frozen; only the $1{\times}1$ gate and the linear head are trained with the BCE route loss. \emph{Shuffled boxes}: each image's box is replaced by another image's box via a within-batch permutation (roll by one), so box statistics are preserved but extents are wrong. \emph{Residual route}: $H_e=H\odot(1+\eta M)$ with $\eta{=}0.5$, trained with the same balanced-BCE route loss and $\lambda_{\mathrm{route}}$, so an ungated path always reaches the classifier. Checkpoints are selected by validation accuracy; test labels never enter selection.

\begin{table}[htb]
\centering
\small
\setlength{\tabcolsep}{2pt}
\begin{tabular}{lrrrrrr}
\toprule
Variant & Acc & ECE15 & EIB & PG & MaxBoxV2 & Gain ret. \\
\midrule
CE & $.749$ & $.017$ & $.788$ & $.966$ & $.584{\pm}.001$ & $0\%$ \\
Frozen backbone & $.731$ & $.026$ & $.794$ & $.976$ & $.597{\pm}.000$ & $10\%$ \\
Shuffled boxes & $.734$ & $.018$ & $.829$ & $.992$ & $.619{\pm}.003$ & $27\%$ \\
Residual-BCE & $.755$ & $.019$ & $.814$ & $.987$ & $.714{\pm}.003$ & $99\%$ \\
BCE-BGER & $.748$ & $.016$ & $.840$ & $.995$ & $.715{\pm}.005$ & $100\%$ \\
\bottomrule
\end{tabular}
\caption{BCE-BGER mechanism controls (CUB full test, three seeds, mean values; MaxBoxAccV2 with seed std). ``Gain ret.'' = fraction of the CE$\to$BCE-BGER MaxBoxAccV2 gain retained; descriptive, and the interventions do not compose additively. HCU (confidence $\ge.9$, EIB $<.5$): CE $1.24{\pm}.02\%$, frozen $0.67{\pm}.17\%$, shuffled $0.50{\pm}.11\%$, BCE-BGER $0.06{\pm}.01\%$. Routing increment vs.\ no-routing: $+.032$ (per-seed $.034/.025/.036$; $95\%$ $t$-interval $[.018,.046]$; no accuracy cost). Exclusivity is not load-bearing: residual $.714{\pm}.003$ vs.\ exclusive $.715{\pm}.005$ ($\Delta{=}{+}.0005$, $[-.008,.009]$); the weaker Energy objective shows a small exclusive gain ($+.021$). Source: \texttt{routing\_increment\_seed\_analysis.csv}.}
\label{tab:bce-controls-full}
\end{table}

\section{Complete Chest-X-ray Metrics}
\label{app:medical-full}

Table~\ref{tab:medical-full} gives the full metric set for the main-text lesion-grounding comparison (Tab.~\ref{tab:ext-datasets}), including Pointing Game and AdaECE, which the main table omits for space. All rows are three seeds, DenseNet-121, positive-finding Grad-CAM on positive boxed studies. The Energy-route results for these datasets are separate and appear in Tab.~\ref{tab:vindr-medical} and Tab.~\ref{tab:rsna-medical}.

\begin{table}[htb]
\centering
\small
\setlength{\tabcolsep}{2pt}
\begin{tabular}{llrrr}
\toprule
Data & Method & AdaECE $\downarrow$ & PG $\uparrow$ & MaxBoxV2 $\uparrow$ \\
\midrule
VinDr & CE & $.136{\pm}.014$ & $.226{\pm}.015$ & $.142{\pm}.015$ \\
 & \textit{CAM-Align} & $\mathbf{.083}{\pm}.011$ & $.550{\pm}.005$ & $.255{\pm}.044$ \\
 & BCE-BGER & $.125{\pm}.020$ & $.404{\pm}.032$ & $.209{\pm}.011$ \\
 & \textit{CAM-BCE} & $.102{\pm}.031$ & $\mathbf{.563}{\pm}.014$ & $\mathbf{.361}{\pm}.002$ \\
\midrule
RSNA & CE & $.027{\pm}.007$ & $.411{\pm}.157$ & $.200{\pm}.066$ \\
 & \textit{CAM-Align} & $.025{\pm}.006$ & $.777{\pm}.013$ & $.361{\pm}.018$ \\
 & BCE-BGER & $\mathbf{.013}{\pm}.010$ & $.713{\pm}.017$ & $\mathbf{.427}{\pm}.029$ \\
 & \textit{CAM-BCE} & $\mathbf{.013}{\pm}.005$ & $\mathbf{.782}{\pm}.019$ & $.420{\pm}.015$ \\
\bottomrule
\end{tabular}
\caption{Complete chest-X-ray metrics for the main-text comparison (three seeds, mean$\pm$std). Italics denote direct CAM-proxy supervision. Accuracy, ECE15, and EIB appear in main Tab.~\ref{tab:ext-datasets}.}
\label{tab:medical-full}
\end{table}

\section{RSNA Qualitative Selection Rule}
\label{app:rsna-selection}

The chest-X-ray example in Fig.~\ref{fig:method} is chosen by a rule fixed before inspecting any map. Over all $1{,}422$ positive boxed RSNA test studies we score every case with the seed-0 CE and BCE-BGER models and mark a case \emph{eligible} if (i)~both models classify it correctly ($p\ge.5$), (ii)~it has a single radiologist box, and (iii)~its box area ratio lies in $[.02,.25]$, the typical lesion range (this excludes degenerate near-empty and near-global boxes). $491$ of $1{,}422$ cases are eligible. We rank the eligible cases by $\Delta\text{EIB}=\text{EIB}_{\text{BCE-BGER}}-\text{EIB}_{\text{CE}}$ and export the \textbf{median} case (rank $245$ of $491$) rather than the largest improvement, so the figure shows a typical rather than a best-case transfer: the exported study has $p_{\mathrm{CE}}{=}0.98$, $p_{\mathrm{BCE}}{=}1.00$, and EIB $.235\!\to\!.397$. For reference, the largest-$\Delta$EIB eligible case reaches $.181\!\to\!.640$. The released manifest records every case's eligibility, per-case metrics, and rank, so both the rule and the ranking are auditable.

\section{Accuracy--Localization Frontier Protocol}
\label{app:frontier}

Both families sweep the route/proxy weight under the same validation-only protocol: seed-0 validation accuracy must stay within $1$ percentage point of the matched CE path, the selected weight maximizes each family's own supervised validation objective (never the audited Grad-CAM), and the full test set is evaluated once per sweep point. \textbf{Basis of the Pareto statement}: sweep points are seed-0; each family's validation-selected operating point ($\lambda{=}2$ for both) is additionally trained with three seeds and matches its seed-0 sweep value (gate $.717$ vs.\ $.715{\pm}.005$; proxy $.771$ vs.\ $.769{\pm}.008$). The gate's selection grid was $\lambda\in\{0.5,1,2\}$; $\lambda\in\{0.25,4\}$ were added afterwards to trace the frontier ends and were never selection candidates.

\begin{table}[htb]
\centering
\small
\setlength{\tabcolsep}{2.2pt}
\begin{tabular}{llrrrr}
\toprule
Family & $\lambda$ & Val Acc & Sel. & Test Acc & Test V2 \\
\midrule
Gate (BCE) & $0.25$ & --- & ends & $.756$ & $.654$ \\
 & $0.5$ & $.722$ & & $.741$ & $.670$ \\
 & $1$ & $.737$ & & $.753$ & $.699$ \\
 & $2$ & $.727$ & $\checkmark$ & $.747$ & $.717$ \\
 & $4$ & --- & ends & $.742$ & $.716$ \\
\midrule
Proxy (Dice) & $.125$ & $.745$ & & $.744$ & $.594$ \\
 & $.25$ & $.732$ & & $.745$ & $.604$ \\
 & $.5$ & $.740$ & & $.744$ & $.620$ \\
 & $1$ & $.738$ & & $.745$ & $.644$ \\
 & $2$ & $.745$ & $\checkmark$ & $.748$ & $.690$ \\
\midrule
Proxy (BCE) & $.0156$ & $.737$ & & $.748$ & $.599$ \\
 & $.0312$ & $.738$ & & $.749$ & $.610$ \\
 & $.0625$ & $.732$ & & $.743$ & $.634$ \\
 & $.125$ & $.728$ & infeas. & $.742$ & $.650$ \\
 & $.25$ & $.742$ & & $.749$ & $.671$ \\
 & $.5$ & $.737$ & & $.748$ & $.710$ \\
 & $1$ & $.742$ & & $.745$ & $.740$ \\
 & $2$ & $.738$ & $\checkmark$ & $.739$ & $.771$ \\
\bottomrule
\end{tabular}
\caption{Frontier sweep protocol (CUB, seed 0 per point; test used once per point). ``Sel.'' marks each family's validation-selected weight, confirmed with three seeds in the main table; ``ends'' marks post-hoc frontier endpoints outside the selection grid; ``infeas.'' failed the validation-accuracy constraint.}
\label{tab:frontier-protocol}
\end{table}

\section{Paired Bootstrap Procedure and Pairwise Intervals}
\label{app:bootstrap}

For each method pair we resample the $5{,}794$ shared test-image IDs with replacement ($10{,}000$ resamples). MaxBoxAccV2 is a max-over-thresholds dataset statistic, so each resample recomputes it from scratch: per method and seed, at each IoU $\delta\in\{.3,.5,.7\}$, the maximum over the 21 thresholds of the resampled hit fraction is taken, averaged over $\delta$ and the three seeds, and the paired difference is formed within the resample. Accuracy differences resample per-image correctness. Intervals are percentile $95\%$ CIs (Tab.~\ref{tab:pairwise-ci}); they capture image-sampling uncertainty, not training-seed variance, which Tab.~\ref{tab:perseed-separation} addresses.

\begin{table}[htb]
\centering
\small
\setlength{\tabcolsep}{2pt}
\begin{tabular}{llrrc}
\toprule
Metric & Comparison & $\Delta$ & $95\%$ CI & $\ne 0$ \\
\midrule
MaxBoxV2 & vs.\ CE & $+.130$ & $[.126,.134]$ & yes \\
 & vs.\ Dice & $+.024$ & $[.022,.027]$ & yes \\
 & vs.\ Energy & $+.057$ & $[.053,.059]$ & yes \\
 & vs.\ CAM-Align & $+.064$ & $[.060,.067]$ & yes \\
 & vs.\ CAM-BCE & $-.054$ & $[-.058,-.051]$ & yes \\
 & Dice vs.\ Energy & $+.032$ & $[.029,.035]$ & yes \\
\midrule
Accuracy & vs.\ CE & $-.001$ & $[-.007,.005]$ & no \\
 & vs.\ Dice & $-.002$ & $[-.006,.003]$ & no \\
 & vs.\ Energy & $-.002$ & $[-.007,.004]$ & no \\
 & vs.\ CAM-Align & $+.000$ & $[-.005,.006]$ & no \\
 & vs.\ CAM-BCE & $+.011$ & $[.005,.016]$ & yes \\
 & Dice vs.\ Energy & $-.000$ & $[-.006,.005]$ & no \\
\bottomrule
\end{tabular}
\caption{Paired image-level bootstrap for the CUB comparison (full test, $10{,}000$ resamples, three-seed means within each resample). Unlabeled rows are BCE-BGER $-$ (method).}
\label{tab:pairwise-ci}
\end{table}

\section{Additional Energy-BGER Diagnostics}
\label{app:energy-diagnostics}

These experiments examine objective-specific behavior of Energy-BGER---component and topology ablations, the route-loss dose response, architecture diagnostics, temperature scaling, natural-image extensions, and medical results---and are reported separately from the BCE-BGER results in the main paper.

\paragraph{Equal-annotation ladder (Energy recipe).}
This ladder was run under the Energy recipe and its evaluator, whose CE reference reads $.588$ rather than the $.584$ of Tab.~\ref{tab:bger-cub-external} (these runs predate the current evaluator build: same $N{=}5{,}794$ images and threshold grid, different map-normalization build; the two agree to $.004$ and no conclusion depends on the difference). Every supervised row receives the identical GT box. Feature routing produces the largest shift among the tested equal-annotation non-proxy controls; input-space delivery and an ABN-style attention branch move the audited map only slightly.

\begin{table*}[tb]
\centering
\small
\setlength{\tabcolsep}{5pt}
\begin{tabular}{lllrrrr}
\toprule
Method & Box use & Route type & Acc & ECE & EIB & MaxBoxV2 \\
\midrule
CE & none & --- & $.749{\pm}.005$ & $.016{\pm}.004$ & $.790{\pm}.001$ & $.588{\pm}.002$ \\
\midrule
Bg-blackout & input mask & input-space & $.723{\pm}.004$ & $.021{\pm}.004$ & $.788{\pm}.007$ & $.595{\pm}.008$ \\
Bg-random & input noise & input-space & $.680{\pm}.004$ & $.039{\pm}.005$ & $.789{\pm}.002$ & $.593{\pm}.009$ \\
Object-crop & input crop & input-space & $.618{\pm}.013$ & $.017{\pm}.006$ & $.816{\pm}.003$ & $.592{\pm}.003$ \\
\midrule
ABN-box & feature & attention branch & $.738{\pm}.004$ & $.024{\pm}.004$ & $.807{\pm}.001$ & $.606{\pm}.006$ \\
w/o box loss & none & exclusive route & $.748{\pm}.003$ & $.018{\pm}.002$ & $.800{\pm}.000$ & $.585{\pm}.001$ \\
\midrule
Energy-BGER & feature & exclusive route & $.750{\pm}.001$ & $.015{\pm}.002$ & $.861{\pm}.001$ & $.668{\pm}.001$ \\
\bottomrule
\end{tabular}
\caption{Equal-annotation controls under the original Energy recipe (CUB, three seeds; this protocol's evaluator reads CE at $.588$ vs.\ $.584$ in Tab.~\ref{tab:bger-cub-external}, see text). Feature routing produces the largest shift among the tested equal-annotation non-proxy controls.}
\label{tab:annotation-ladder}
\end{table*}

\paragraph{Full-test bootstrap uncertainty (Energy route).}
For the original Energy-BGER result, a seed-0 bootstrap over all $5{,}794$ CUB test images ($10{,}000$ resamples) gives CE $0.585$ with $95\%$ CI $[0.581,0.593]$ and Energy-BGER $0.660$ with $[0.655,0.666]$.

\begin{table*}[tb]
\centering
\small
\begin{tabular}{lccccc}
\toprule
Variant & Acc & ECE & Grad-CAM EIB & Grad-CAM MaxBoxAccV2 & Gate EIB \\
\midrule
Energy-BGER (full)     & 0.748 & 0.015 & 0.862 & \textbf{0.667} & 0.796 \\
\;\;w/o box loss       & 0.748 & 0.018 & 0.800 & 0.585 & 0.487 \\
\;\;w/o area loss      & 0.748 & 0.015 & 0.858 & 0.670 & 0.788 \\
\;\;residual routing   & 0.751 & 0.034 & 0.815 & 0.645 & 0.822 \\
\;\;frozen backbone    & 0.732 & 0.034 & 0.808 & 0.603 & 0.722 \\
\;\;shuffled-box control & 0.741 & 0.018 & 0.855 & 0.608 & 0.795 \\
\midrule
$\rho=0.10$            & 0.749 & 0.018 & 0.863 & 0.667 & 0.801 \\
$\rho=0.25$ (selected) & 0.748 & 0.015 & 0.862 & 0.667 & 0.796 \\
$\rho=0.50$            & 0.751 & 0.017 & 0.859 & 0.669 & 0.790 \\
\bottomrule
\end{tabular}
\caption{Component and topology controls for the original Energy objective on CUB. These results diagnose routing design and should not be read as BCE-BGER ablations.}
\label{tab:ablation}
\end{table*}

\begin{figure*}[tb]
    \centering
    \includegraphics[width=0.8\textwidth]{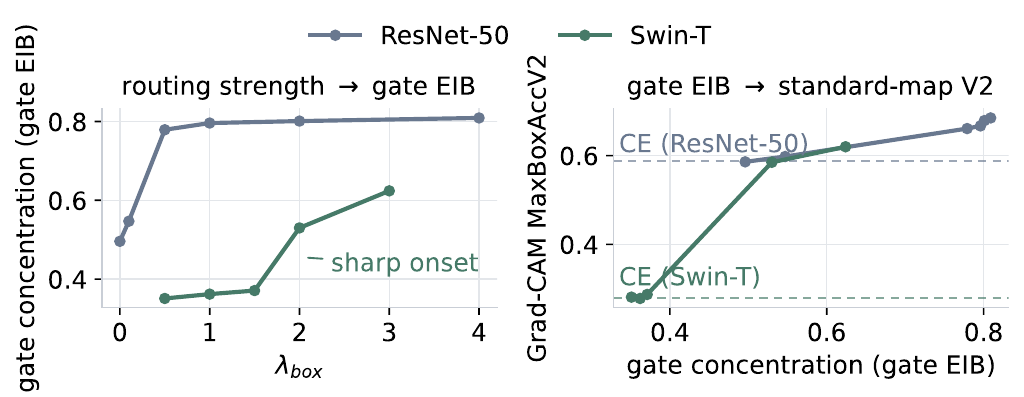}
    \caption{Route-loss dose response on CUB (seed-0 sweep points; Swin $\lambda_{\mathrm{box}} \in \{1, 2\}$ are three-seed means). Gate concentration rises smoothly on ResNet-50 and sharply between $\lambda_{\mathrm{box}}{=}1.5$ and $2$ on Swin-T; dashed lines mark each architecture's CE Grad-CAM baseline.}
    \label{fig:dose-response}
\end{figure*}

\begin{table*}[tb]
\centering
\small
\begin{tabular}{lrrrr}
\toprule
Method & Acc $\uparrow$ & ECE15 $\downarrow$ & EIB $\uparrow$ & MaxBoxAccV2 $\uparrow$ \\
\midrule
\multicolumn{5}{l}{\emph{Swin-T transformer}}\\
CE & $.786{\pm}.005$ & $.015{\pm}.004$ & $.360{\pm}.003$ & $.280{\pm}.002$ \\
CAM-Align & $\mathbf{.791}{\pm}.002$ & $\mathbf{.013}{\pm}.002$ & $.413{\pm}.003$ & $.343{\pm}.010$ \\
Energy-BGER ($\lambda_{\mathrm{box}}{=}1.0$) & $.783{\pm}.002$ & $.014{\pm}.003$ & $.372{\pm}.005$ & $.278{\pm}.003$ \\
Energy-BGER ($\lambda_{\mathrm{box}}{=}2.0$) & $.789{\pm}.005$ & $.145{\pm}.006$ & $\mathbf{.567}{\pm}.008$ & $\mathbf{.585}{\pm}.021$ \\
\quad + temperature scaling & $.789{\pm}.005$ & $.018{\pm}.004$ & $.567{\pm}.008$ & $.585{\pm}.021$ \\
\midrule
\multicolumn{5}{l}{\emph{ConvNeXt-T CNN}}\\
CE & $\mathbf{.812}{\pm}.002$ & $.142{\pm}.018$ & $.524{\pm}.020$ & $.291{\pm}.010$ \\
CAM-Align & $.809{\pm}.005$ & $\mathbf{.132}{\pm}.009$ & $\mathbf{.706}{\pm}.003$ & $\mathbf{.370}{\pm}.005$ \\
Energy-BGER ($\lambda_{\mathrm{box}}{=}1.0$) & $.810{\pm}.004$ & $.141{\pm}.017$ & $.437{\pm}.025$ & $.298{\pm}.020$ \\
Energy-BGER ($\lambda_{\mathrm{box}}{=}4.0$) & $.798{\pm}.015$ & $.161{\pm}.022$ & $.516{\pm}.018$ & $.333{\pm}.005$ \\
\bottomrule
\end{tabular}
\caption{Architecture diagnostics on CUB (three seeds, Energy route). Stronger Energy routing raises the standard audited map on Swin-T, but audit validity is not established for these Energy-route checkpoints (the BCE-BGER detach audit in Tab.~\ref{tab:detach-arch} shows Swin standard-map gains can be gate-gradient imprinting); ConvNeXt-T shows a smaller gain with an accuracy cost. Operating points are selected on validation data (Fig.~\ref{fig:dose-response}, Tab.~\ref{tab:convnext-sweep}).}
\label{tab:swin-generality}
\end{table*}

\paragraph{Temperature scaling: proper scoring and attribution invariance.}
Temperature scaling (TS), fit on validation NLL, changes Swin-T Energy-BGER NLL/Brier from $0.908/0.331$ to $0.794/0.300$ and ECE15 from $.145$ to $.017$. ConvNeXt-T CE already has ECE $.142$; TS reduces every method to $.014$--$.015$, while Energy-BGER $\lambda_{\mathrm{box}}{=}4$ retains an NLL cost over CE ($0.759$ vs.\ $0.698$). Scaling logits by $1/T$ uniformly scales Grad-CAM gradients and cancels under map normalization; the measured maximum map difference is $3.4\times10^{-6}$.

\begin{table}[htb]
\centering
\small
\setlength{\tabcolsep}{4pt}
\setlength{\tabcolsep}{2pt}
\begin{tabular}{rrrrrr}
\toprule
$\lambda_{\mathrm{box}}$ & val Acc & val Gate EIB & test Acc & test EIB & test MaxBoxV2 \\
\midrule
0 (CE) & .778 & --- & .815 & .526 & .290 \\
1 & .777 & .826 & .806 & .438 & .295 \\
2 & .773 & .853 & .802 & .511 & .336 \\
4 & .765 & .883 & .799 & .516 & .338 \\
8 & .770 & .893 & .784 & .562 & .342 \\
\bottomrule
\end{tabular}
\caption{ConvNeXt-T route-loss sweep on CUB (seed 0). $\lambda_{\mathrm{box}}{=}0$ is CE; $\lambda_{\mathrm{box}}{=}4$ is selected from validation gate concentration and accuracy.}
\label{tab:convnext-sweep}
\end{table}

\begin{figure}[htb]
    \centering
    \includegraphics[width=\columnwidth]{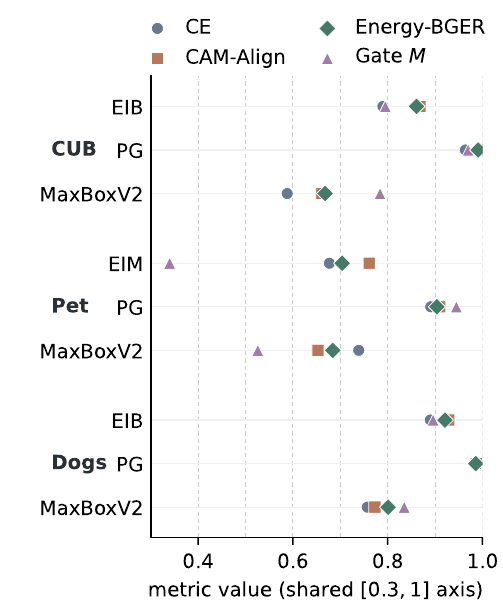}
    \caption{Cross-dataset localization (three-seed means): each row is one dataset--metric, and the four methods are points on a common $[0.3,1]$ axis (no per-axis rescaling), so within-row method comparison is valid and no cross-metric magnitude is implied. Complete estimates are in Tabs.~\ref{tab:bger-cub-external} and \ref{tab:multi}.}
    \label{fig:multi-summary}
\end{figure}

\begin{table*}[tb]
\centering
\small
\setlength{\tabcolsep}{3pt}
\begin{tabular}{lllrrrrr}
\toprule
Dataset & Method & Evidence target & Acc & ECE & EIB/EIM & PG & MaxBoxAccV2 \\
\midrule
Oxford-IIIT Pet & CE & foreground mask & $.930{\pm}.000$ & $.022{\pm}.005$ & $.677{\pm}.002$ & $.891{\pm}.004$ & $\mathbf{.739}{\pm}.002$ \\
Oxford-IIIT Pet & CAM-Align & foreground mask & $.927{\pm}.003$ & $.019{\pm}.002$ & $\mathbf{.761}{\pm}.003$ & $\mathbf{.910}{\pm}.002$ & $.653{\pm}.005$ \\
Oxford-IIIT Pet & Energy-BGER & foreground mask & $\mathbf{.931}{\pm}.000$ & $\mathbf{.016}{\pm}.003$ & $.704{\pm}.006$ & $.904{\pm}.006$ & $.684{\pm}.006$ \\
Oxford-IIIT Pet & Energy gate $M$ & diagnostic gate & -- & -- & $.340{\pm}.001$ & $.945{\pm}.003$ & $.526{\pm}.002$ \\
Stanford Dogs & CE & dog box & $\mathbf{.880}{\pm}.005$ & $.045{\pm}.003$ & $.890{\pm}.000$ & $.984{\pm}.000$ & $.757{\pm}.004$ \\
Stanford Dogs & CAM-Align & dog box & $.875{\pm}.002$ & $\mathbf{.041}{\pm}.003$ & $\mathbf{.929}{\pm}.001$ & $\mathbf{.987}{\pm}.000$ & $.773{\pm}.006$ \\
Stanford Dogs & Energy-BGER & dog box & $.877{\pm}.003$ & $.042{\pm}.001$ & $.921{\pm}.001$ & $.986{\pm}.000$ & $\mathbf{.801}{\pm}.002$ \\
Stanford Dogs & Energy gate $M$ & diagnostic gate & -- & -- & $.896{\pm}.002$ & $.987{\pm}.001$ & $.835{\pm}.001$ \\
\bottomrule
\end{tabular}
\caption{Natural-image extensions with Energy-BGER (three seeds). BCE-BGER Dogs results are reported in the main paper.}
\label{tab:multi}
\end{table*}

\begin{figure*}[tb]
    \centering
    \includegraphics[width=0.72\textwidth]{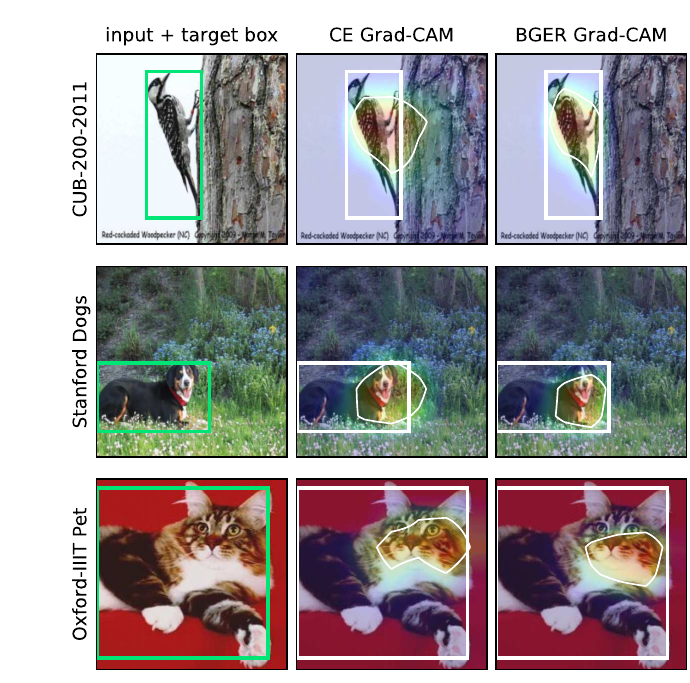}
    \caption{Representative route-to-CAM examples for Energy-BGER on CUB, Stanford Dogs, and Oxford-IIIT Pet. Columns show the input with the target box, the CE predicted-class Grad-CAM, and the Energy-BGER predicted-class Grad-CAM (seed 0); the white contour marks the map's $0.5$ level. Each row is the largest CE$\to$BGER EIB-gain case found in a fixed-stride scan of the test split, so Energy-BGER concentrates the audited map inside the target relative to CE.}
    \label{fig:qual}
\end{figure*}

\begin{table*}[tb]
\centering
\small
\setlength{\tabcolsep}{3pt}
\begin{minipage}[t]{0.5\textwidth}
\centering
\begin{tabular}{lrrrr}
\toprule
Method & AUROC $\uparrow$ & AUPRC $\uparrow$ & Acc $\uparrow$ & ECE15 $\downarrow$ \\
\midrule
CE & $.851{\pm}.021$ & $.768{\pm}.027$ & $.808{\pm}.008$ & $.138{\pm}.012$ \\
CAM-Align & $\mathbf{.878}{\pm}.027$ & $\mathbf{.805}{\pm}.028$ & $.828{\pm}.007$ & $\mathbf{.085}{\pm}.013$ \\
Energy-BGER & $.867{\pm}.025$ & $\mathbf{.805}{\pm}.025$ & $\mathbf{.835}{\pm}.011$ & $.104{\pm}.022$ \\
\bottomrule
\end{tabular}
\end{minipage}\hfill
\begin{minipage}[t]{0.5\textwidth}
\centering
\begin{tabular}{lrrr}
\toprule
Map & EIB/EIM $\uparrow$ & PG $\uparrow$ & MaxBoxV2 $\uparrow$ \\
\midrule
CE & $.147{\pm}.012$ & $.226{\pm}.015$ & $.142{\pm}.015$ \\
CAM-Align & $.390{\pm}.006$ & $.550{\pm}.005$ & $.255{\pm}.044$ \\
Energy-BGER & $.272{\pm}.030$ & $.416{\pm}.054$ & $.212{\pm}.018$ \\
Energy gate $M$ & $.350{\pm}.023$ & $.494{\pm}.024$ & $.217{\pm}.009$ \\
\bottomrule
\end{tabular}

\end{minipage}
\caption{VinDr-CXR results with Energy-BGER. BCE-BGER and CAM-BCE results are reported in the main paper.}
\label{tab:vindr-medical}
\end{table*}

\begin{table*}[tb]
\centering
\small
\setlength{\tabcolsep}{3pt}
\begin{minipage}[t]{0.5\textwidth}
\centering
\begin{tabular}{lrrrr}
\toprule
Method & AUROC $\uparrow$ & AUPRC $\uparrow$ & Acc $\uparrow$ & ECE15 $\downarrow$ \\
\midrule
CE & $.985{\pm}.002$ & $.983{\pm}.003$ & $.947{\pm}.003$ & $.027{\pm}.007$ \\
CAM-Align & $\mathbf{.986}{\pm}.001$ & $.983{\pm}.001$ & $.946{\pm}.003$ & $.024{\pm}.011$ \\
Energy-BGER & $\mathbf{.986}{\pm}.002$ & $\mathbf{.984}{\pm}.002$ & $\mathbf{.953}{\pm}.003$ & $\mathbf{.021}{\pm}.016$ \\
\bottomrule
\end{tabular}
\end{minipage}\hfill
\begin{minipage}[t]{0.5\textwidth}
\centering
\begin{tabular}{lrrr}
\toprule
Map & EIB/EIM $\uparrow$ & PG $\uparrow$ & MaxBoxV2 $\uparrow$ \\
\midrule
CE & $.189{\pm}.036$ & $.411{\pm}.157$ & $.200{\pm}.066$ \\
CAM-Align & $.565{\pm}.017$ & $.777{\pm}.013$ & $.361{\pm}.018$ \\
Energy-BGER & $.520{\pm}.014$ & $.745{\pm}.037$ & $.305{\pm}.026$ \\
Energy gate $M$ & $.642{\pm}.034$ & $.768{\pm}.035$ & $.264{\pm}.025$ \\
\bottomrule
\end{tabular}
\end{minipage}
\caption{RSNA results with Energy-BGER (three seeds, validation-selected $\rho=0.02$). BCE-BGER and CAM-BCE results are reported in the main paper.}
\label{tab:rsna-medical}
\end{table*}

\paragraph{RSNA subgroup analysis.}
For Energy-BGER, soft-energy and thresholded-IoU changes diverge with target area and multi-box structure (Tab.~\ref{tab:rsna-subgroup}).

\section{COCO-SO Stress Test}

COCO-SO is a seed-0, single-target subset derived from COCO 2017 for testing cluttered scenes. This pilot uses the Energy route.

\begin{table*}[t]
\centering
\begin{minipage}[t]{0.49\textwidth}
\centering
\small
\setlength{\tabcolsep}{2pt}
\begin{tabular}{lrrrr}
\toprule
Subgroup & $n$ & $\Delta$EIB & $\Delta$PG & $\Delta$BestIoU \\
\midrule
all positive boxed & 1422 & $+$.300 & $+$.462 & $+$.072 \\
single-box & 643 & $+$.226 & $+$.311 & $+$.096 \\
multi-box ($\ge$2) & 779 & $+$.361 & $+$.587 & $+$.053 \\
small box (tercile 1) & 474 & $+$.172 & $+$.251 & $+$.082 \\
medium box (tercile 2) & 474 & $+$.308 & $+$.489 & $+$.116 \\
large box (tercile 3) & 474 & $+$.420 & $+$.646 & $+$.020 \\
EIB up, BestIoU down & 449 & $+$.333 & $+$.359 & $-$.138 \\
gate loc., CAM diffuse & 91 & $+$.069 & $+$.440 & $+$.040 \\
\bottomrule
\end{tabular}
\captionof{table}{RSNA subgroup analysis (positive, boxed cases; per-sample Grad-CAM, seed 0). Spatial improvement (CE$\to$BGER) varies with box count and size; on a minority of cases soft evidence (EIB) improves while thresholded box overlap (BestIoU) does not. Box-area terciles at $0.052$/$0.124$ of image area. Controlled extension, not clinical validation.}
\label{tab:rsna-subgroup}
\end{minipage}\hfill
\begin{minipage}[t]{0.48\textwidth}
\centering
\small
\setlength{\tabcolsep}{2pt}
\begin{tabular}{lrrrrr}
\toprule
Method & Acc & ECE & EIM & PG & MaxBoxAccV2 \\
\midrule
CE & \textbf{0.983} & \textbf{0.012} & 0.483 & 0.727 & \textbf{0.658} \\
Energy-BGER & 0.971 & 0.037 & \textbf{0.579} & \textbf{0.740} & 0.627 \\
Energy-BGER $M$ & -- & -- & 0.544 & 0.746 & 0.620 \\
\bottomrule
\end{tabular}
\captionof{table}{COCO-SO seed-0 appendix stress test (Energy route). Energy-BGER improves soft target-mask energy and PG, but not MaxBoxAccV2, accuracy, or calibration.}
\label{tab:coco}
\end{minipage}
\end{table*}

In 352/524 samples, Energy-BGER increases EIM but decreases best box IoU. The class-agnostic route can concentrate soft energy on the target without producing an instance-complete thresholded box.

\section{Exploratory Energy-Route Medical Probes}
\label{sec:new-medical}
Three exploratory Energy-BGER probes cover brain CT \citep{hssayeni2020ich}, mammography \citep{nguyen2023vindrmammo}, and lumbar MRI \citep{rsna2024lumbar} (Tab.~\ref{tab:new-medical}). Soft in-target energy rises on all three; thresholded localization does not improve on Lumbar.

\begin{table*}[t]
\centering
\footnotesize
\setlength{\tabcolsep}{6pt}
\begin{tabular}{llrrrrr}
\toprule
Dataset & Method & AUROC $\uparrow$ & ECE15 $\downarrow$ & EIM $\uparrow$ & PG $\uparrow$ & MaxBoxV2 $\uparrow$ \\
\midrule
\multirow{3}{*}{ICH (brain CT)} & CE & $.834$ & $.072$ & $.018$ & $.030$ & $.034$ \\
 & CAM-Align & $.836$ & $.061$ & $.018$ & $.033$ & $.044$ \\
 & Energy-BGER & $.845$ & $.074$ & $.022$ & $.052$ & $.060$ \\
\midrule
\multirow{3}{*}{VinDr-Mammo} & CE & $.792$ & $.140$ & $.024$ & $.021$ & $.008$ \\
 & CAM-Align & $.779$ & $.107$ & $.090$ & $.160$ & $.053$ \\
 & Energy-BGER & $.772$ & $.109$ & $.059$ & $.076$ & $.029$ \\
\midrule
\multirow{3}{*}{Lumbar (SCS)} & CE & $.874$ & $.084$ & $.057$ & $.308$ & $.163$ \\
 & CAM-Align & $.874$ & $.044$ & $.109$ & $.152$ & $.048$ \\
 & Energy-BGER & $.883$ & $.059$ & $.108$ & $.140$ & $.046$ \\
\bottomrule
\end{tabular}
\caption{Exploratory Energy-BGER medical extensions (three-seed means; DenseNet-121; boxed abnormality versus no finding). Spatial metrics use positive boxed test images.}
\label{tab:new-medical}
\end{table*}

\section{Direct CAM-Supervision Objectives}
\label{app:proxy}

CAM-Align uses the same backbone, recipe, and evidence target as Energy-BGER and differs in the supervised object. From the final feature map $H$ and true-class head weights $w_y$, its proxy is $P = \mathrm{ReLU}\big(\textstyle\sum_c w_{yc} H_c\big)$. The alignment loss applies the inside-region energy objective to $P$:
\begin{equation}
\mathcal{L}_{\mathrm{align}} = 1 - \frac{\sum_u P_u B_u}{\sum_u P_u + \epsilon}, \qquad \mathcal{L} = \mathcal{L}_{\mathrm{cls}} + \lambda_{\mathrm{cam}}\,\mathcal{L}_{\mathrm{align}},
\end{equation}
with $\lambda_{\mathrm{cam}} = 1.0$, selected by the same validation sweep used for Energy-BGER's $\lambda_{\mathrm{box}}$.

The two objective-matched proxies use the same bounded map $Q=\sigma(\sum_c w_{yc}H_c)$, which is the direct-supervision analogue of the gate $M=\sigma(\mathrm{Conv}_{1\times1}(H))$: both are sigmoid maps in $[0,1]^{1\times h\times w}$, so the identical route loss can be applied to either. \emph{CAM-BCE} applies the per-image balanced BCE of Eq.~\ref{eq:bbce} to $Q$ instead of $M$. \emph{CAM-Dice} applies the soft-Dice objective of Dice-BGER to $Q$ instead of $M$, i.e.\ $1-2\sum_u Q_uB_u/(\sum_u Q_u+\sum_u B_u+\epsilon)$. Both use the same evidence target $B$, backbone, recipe, seeds, and evaluator as the routes, and both take $\lambda_{\mathrm{route}}=2$, selected on seed-0 validation by the same firewalled rule as the gates (maximize the method's own supervised validation object subject to validation accuracy within $1$ point of the matched CE path; the audited Grad-CAM never enters selection). Their weight sweeps and selected points are in Tab.~\ref{tab:frontier-protocol}. The proxy and route arms therefore match the evidence target, objective family, backbone, training recipe, seeds, selection rule, and evaluator. They differ in the supervised object---an attribution-family proxy close to the audited map versus an internal gate---and, by construction, in whether that object lies on the classification path ($\mathrm{GAP}(H)$ for the proxies, $\mathrm{GAP}(M{\odot}H)$ for the routes). The auxiliary no-routing control (App.~\ref{app:bce-controls}) isolates the routing contribution separately.

\section{Gate-Gradient Effects on Grad-CAM}

The frozen-backbone control bounds the non-representational part of the transfer empirically; here we characterize it analytically for the ResNet instantiation, where the audited layer $H$ directly feeds the gate. With a GAP--linear head on $H_e = M \odot H$, $M = \sigma(g)$, $g = v^\top H + b$ (the $1{\times}1$ gate with weight $v \in \mathbb{R}^C$ and bias $b$), the score for class $k$ is $s_k = \frac{1}{hw}\sum_u w_k^\top (M_u H_u) + c_k$. Differentiating through both paths (the routed features and the gate itself; the bias terms drop out),
\begin{equation}
\frac{\partial s_k}{\partial H_{c,u}} \;=\; \frac{1}{hw}\Big[\, w_{kc}\, M_u \;+\; \sigma'(g_u)\, (w_k^\top H_u)\, v_c \Big],
\end{equation}
so the Grad-CAM channel weights (spatial GAP of the gradient) decompose \emph{exactly} as
\begin{equation}
\boldsymbol{\alpha}_k \;=\; \mu\, w_k \;+\; \beta_k\, v,
\label{eq:alpha-decomp}
\end{equation}
with $\mu = \sum_u M_u/(hw)^2 > 0$ and $\beta_k = \sum_u \sigma'(g_u)(w_k^\top H_u)/(hw)^2$.
The direct mechanical effect is confined to a positive rescaling along $w_k$, which vanishes under map normalization, and the image-independent direction $v$. On the full CUB test set, the projection residual onto $\mathrm{span}\{w_k,v\}$ is $1.4\times10^{-7}$, and audited Grad-CAM correlates $0.999\pm0.000$ with the pre-gate head-weight CAM. Together with the frozen-backbone control, this identifies changes in $H$ as the dominant source of route-to-CAM transfer.

\paragraph{Empirical detach test across architectures.}
We test the decomposition's consequence directly on every backbone: recompute the audited predicted-class Grad-CAM with the gate gradient $\partial M/\partial H$ detached (forward gate values preserved) and compare to the standard map (Tab.~\ref{tab:detach-arch}). This is the complete per-backbone evidence behind the main-paper architecture table. On ResNet-50 the detached map is essentially identical to the standard map (mean per-image correlation $.9999$, MaxBoxAccV2 $.715\!\to\!.712$). On Swin-T it collapses to $.256$---below CE---with anti-correlated maps ($-.16$): the standard routed map fails the detach-based gate-imprinting check, so it is not evidence of representation transfer. On ConvNeXt-T the check itself passes ($.253\!\to\!.255$, $r{=}.974$) while routing lowers localization below CE: a genuine method failure under this backbone, distinct from Swin-T's audit failure.

\begin{table}[htb]
\centering
\small
\setlength{\tabcolsep}{2.6pt}
\begin{tabular}{lrrrrr}
\toprule
Backbone & CE & Aux & BGER & BGER det. & corr \\
\midrule
ResNet-50 & $.584$ & $.683$ & $.715{\pm}.005$ & $.712{\pm}.006$ & $.9999$ \\
Swin-T & $.280$ & $.317$ & $.896{\pm}.000$ & $.256{\pm}.001$ & $-.16{\pm}.04$ \\
ConvNeXt-T & $.291$ & $.288$ & $.253{\pm}.002$ & $.255{\pm}.002$ & $.974{\pm}.004$ \\
\bottomrule
\end{tabular}
\caption{Complete architecture mechanism arms with detach audit (CUB, three seeds, full test, MaxBoxAccV2): CE, auxiliary BCE without routing (Aux), routed BCE-BGER, BCE-BGER with the gate gradient detached, and the mean per-image standard-vs-detached map correlation. The detach-based gate-imprinting check passes on ResNet-50 and ConvNeXt-T but fails on Swin-T, and only ResNet-50 combines a passing check with a positive gain. DenseNet-121 medical audits appear separately in Tab.~\ref{tab:detach-densenet}; source files and reproduction commands are listed in the submitted artifact manifest.}
\label{tab:detach-arch}
\end{table}

\paragraph{DenseNet-121 medical detach audit.}
The chest-X-ray models use DenseNet-121 with a GAP--linear positive-abnormality head. We run the detach test on the reported BCE-BGER checkpoints (positive boxed test images, three seeds; Tab.~\ref{tab:detach-densenet}). Both datasets remain substantially stable under detachment: standard-versus-detached map correlations are $.957$ (RSNA) and $.940$ (VinDr), while $94\%$ and $78\%$ of the respective localization gains are retained, and the detached MaxBoxAccV2 stays well above CE---no collapse, unlike Swin-T, which licenses treating the DenseNet-121 medical maps as valid audits.

\begin{table}[htb]
\centering
\small
\setlength{\tabcolsep}{4pt}
\begin{tabular}{lrrrr}
\toprule
Data & std V2 & det V2 & corr & retained \\
\midrule
RSNA & $.427$ & $.413$ & $.957$ & $94\%$ \\
VinDr & $.208$ & $.194$ & $.940$ & $78\%$ \\
\bottomrule
\end{tabular}
\caption{DenseNet-121 detach audit on the medical BCE-BGER checkpoints (positive boxed test images, three seeds): standard vs.\ detached MaxBoxAccV2, mean per-image map correlation, and the fraction of the CE$\to$BCE-BGER gain retained after detachment. Both models remain substantially stable under detachment; VinDr's lower retention reflects its small absolute gain ($.208{-}.142$), while the map correlation ($.94$) shows the map itself is preserved. The ``std V2'' column is recomputed by this audit and agrees with Tab.~\ref{tab:ext-datasets} to within rounding (VinDr $.2082$ vs.\ $.2089$; RSNA $.4268$ vs.\ $.4269$). Source files and reproduction commands are listed in the submitted artifact manifest.}
\label{tab:detach-densenet}
\end{table}

\paragraph{Beyond ResNet (Energy-route frozen-backbone).} As a separate diagnostic on the earlier Energy-route models, Swin-T and ConvNeXt-T frozen-backbone controls recover at most a fifth of the Energy-route standard-map gain: Swin-T reaches $0.343\pm0.019$ ($21\%$ of the $0.280\to0.585$ Energy-route gap), and ConvNeXt-T reaches $0.254\pm0.000$, below CE ($0.291$).

\section{Region-Restricted Corruption and HCU Protocol}
\label{app:corruption}

The double dissociation in Fig.~\ref{fig:dissociation} uses the BCE-BGER checkpoints (seeds 0--2) and CE on the fixed 1000-image CUB evaluation subset. Each image is corrupted either only inside the ground-truth box (\emph{object}) or only outside it (\emph{background}), by compositing a corrupted copy with the original through the box mask in denormalized pixel space. Two corruption types are used: additive Gaussian noise with $\sigma \in \{0.10, 0.25, 0.50\}$ and Gaussian blur with kernel/std $\{(9,3), (21,7), (41,15)\}$ pixels at $224^2$ resolution, giving severities 1--3; each condition uses a fixed random generator so all methods see identical corruptions, and accuracy is averaged over the two corruption types per seed. Severity-3 means: background-corrupt CE $.711{\pm}.024$ vs.\ BCE-BGER $.720{\pm}.027$; object-corrupt CE $.048{\pm}.006$ vs.\ BCE-BGER $.021{\pm}.006$ (per-condition values released with the audit records). The same protocol run under the original Energy objective (CE, CAM-Align, Energy-BGER) shows the same dissociation direction and is retained below.

\paragraph{Full-test HCU.} We compute HCU on all $5{,}794$ CUB test images as the fraction of predictions with confidence $\ge 0.9$ whose Grad-CAM EIB is $< 0.5$. Values: CE $1.24{\pm}.02\%$, BCE-BGER $0.06{\pm}.01\%$ (frozen $0.67{\pm}.17\%$, shuffled $0.50{\pm}.11\%$; Tab.~\ref{tab:bce-controls-full}).

\section{CAM Concentration vs.\ Shortcut Robustness}
\label{app:spurious}
This controlled counterexample supports the limitation that improved CAM concentration does not guarantee background-shortcut robustness. We build a split-disjoint spurious-context version of Oxford-IIIT Pet in which foreground species is spuriously correlated with background at training time, with foreground and background pools drawn from disjoint image sets so the background texture cannot leak object identity (leakage controls in the released records). Under the balanced-BCE gate (Gate-BCE) and its objective-matched CAM proxy (CAM-BCE), the audited maps concentrate strongly (Gate-BCE audit EIB $.348$, MaxBoxAccV2 $.235$; CAM-BCE $.615$/$.503$). Yet on a foreground-only balanced test the same models are at chance ($.500$--$.52$ accuracy), and a counterfactual background swap flips roughly three quarters of predictions (Gate-BCE flip rate $.74$, CAM-BCE $.74$). Concentrated post-hoc maps therefore coexist with strong background dependence: CAM concentration is not evidence of shortcut robustness. Localization, counterfactual, foreground-only, leakage-control, and paired-bootstrap records are released under \texttt{spurious\_pet\_disjoint/}.

\begin{figure*}[tb]
    \centering
    \includegraphics[width=0.8\textwidth]{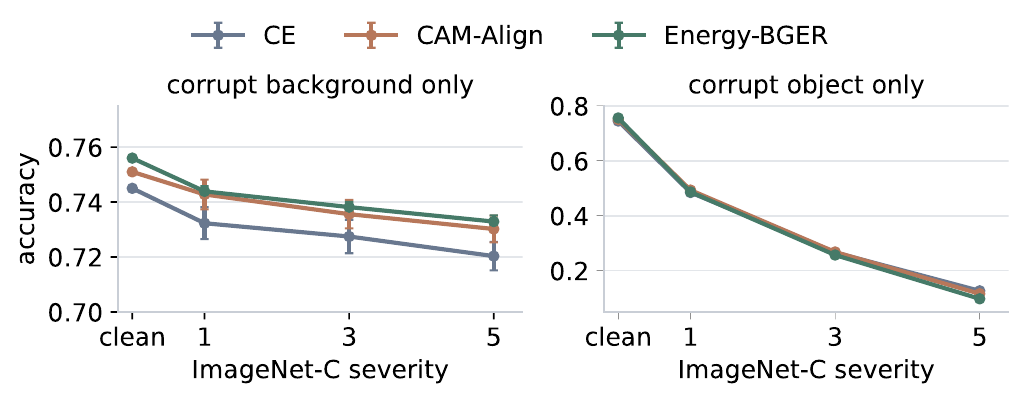}
    \caption{Original Energy-objective region-restricted diagnostic under the standard ImageNet-C suite. Each point is the mean over 10 corruption types and 3 seeds; error bars are the standard deviation across seeds. Energy-BGER is relatively less sensitive to background corruption and more sensitive to object corruption.}
    \label{fig:inc-dissociation}
\end{figure*}

\paragraph{Standard ImageNet-C corruption suite.}
To confirm the dissociation is not specific to our noise/blur choices, we repeat the original Energy protocol with ten standard ImageNet-C corruption types \citep{hendrycks2019imagenetc} at their original $224^2$ severity constants (severities 1/3/5): Gaussian noise, shot noise, impulse noise, defocus blur, motion blur, fog, brightness, contrast, pixelate, and JPEG compression. These are asset-free re-implementations of the reference code; frost and snow (external image assets) and glass/zoom blur and elastic transform (per-pixel loops, computationally prohibitive at this scale) are excluded. Fig.~\ref{fig:inc-dissociation} shows the same double-dissociation direction: under background corruption Energy-BGER degrades least (severity-5 accuracy $0.733$ vs.\ CAM-Align $0.730$ vs.\ CE $0.720$), and under object corruption it degrades most ($0.098$ vs.\ $0.116$ vs.\ $0.127$).

\paragraph{HCU threshold sensitivity (Energy-BGER).}
Tab.~\ref{tab:hcu-sensitivity} is the original Energy-route threshold sweep on the fixed $N=1{,}000$ records; the CE-vs-routed ordering persists across all tested confidence and EIB thresholds, which also supports the full-test statistic above.

\begin{table}[htb]
\centering
\small
\begin{tabular}{llrrr}
\toprule
$\tau$ & $\theta$ & CE & CAM-Align & Energy-BGER \\
\midrule
.7  & .3 & $.07{\pm}.05$ & $.00{\pm}.00$ & $.00{\pm}.00$ \\
.7  & .5 & $2.20{\pm}.36$ & $.07{\pm}.05$ & $.13{\pm}.05$ \\
.7  & .7 & $15.40{\pm}.08$ & $4.07{\pm}.17$ & $4.37{\pm}.17$ \\
.8  & .3 & $.07{\pm}.05$ & $.00{\pm}.00$ & $.00{\pm}.00$ \\
.8  & .5 & $1.50{\pm}.24$ & $.07{\pm}.05$ & $.10{\pm}.00$ \\
.8  & .7 & $12.60{\pm}.51$ & $3.20{\pm}.36$ & $3.53{\pm}.12$ \\
.9  & .3 & $.03{\pm}.05$ & $.00{\pm}.00$ & $.00{\pm}.00$ \\
.9  & .5 & $1.10{\pm}.08$ & $.07{\pm}.05$ & $.07{\pm}.05$ \\
.9  & .7 & $8.67{\pm}.21$ & $2.00{\pm}.08$ & $2.37{\pm}.19$ \\
.95 & .3 & $.03{\pm}.05$ & $.00{\pm}.00$ & $.00{\pm}.00$ \\
.95 & .5 & $.77{\pm}.05$ & $.03{\pm}.05$ & $.00{\pm}.00$ \\
.95 & .7 & $6.10{\pm}.29$ & $1.10{\pm}.08$ & $1.53{\pm}.19$ \\
\bottomrule
\end{tabular}
\caption{Threshold sensitivity of the original Energy-BGER HCU diagnostic on CUB ($N=1{,}000$, three seeds).}
\label{tab:hcu-sensitivity}
\end{table}

\section{Attribution-Free Region Diagnostics}

Blurring the annotated region or its complement changes Energy-BGER and CAM-Align confidence in the same direction as the spatial audit (Tab.~\ref{tab:cam-free}). These confidence diagnostics do not establish causal reliance or distributional robustness.

\begin{table*}[tb]
\centering
\small
\begin{tabular}{lrrrr}
\toprule
Method & RemoveBoxDrop $\uparrow$ & KeepBgConf $\downarrow$ & EvidenceGap $\uparrow$ & Del-AUC$_n$ $\downarrow$ \\
\midrule
CE & $.746{\pm}.005$ & $.012{\pm}.000$ & $.599{\pm}.001$ & $.082{\pm}.001$ \\
CAM-Align & $\mathbf{.759}{\pm}.006$ & $\mathbf{.005}{\pm}.000$ & $\mathbf{.624}{\pm}.002$ & $\mathbf{.065}{\pm}.001$ \\
Energy-BGER & $.752{\pm}.003$ & $\mathbf{.005}{\pm}.000$ & $.615{\pm}.002$ & $.067{\pm}.000$ \\
\bottomrule
\end{tabular}
\caption{Attribution-free region-restricted confidence diagnostics for the original Energy objective on CUB (three seeds). Deletion-AUC is attribution-guided and listed only as a secondary measure.}
\label{tab:cam-free}
\end{table*}

\section{Gate Transfer and Design Diagnostics}

\begin{figure*}[tb]
    \centering
    \includegraphics[width=\textwidth,height=0.55\textheight,keepaspectratio]{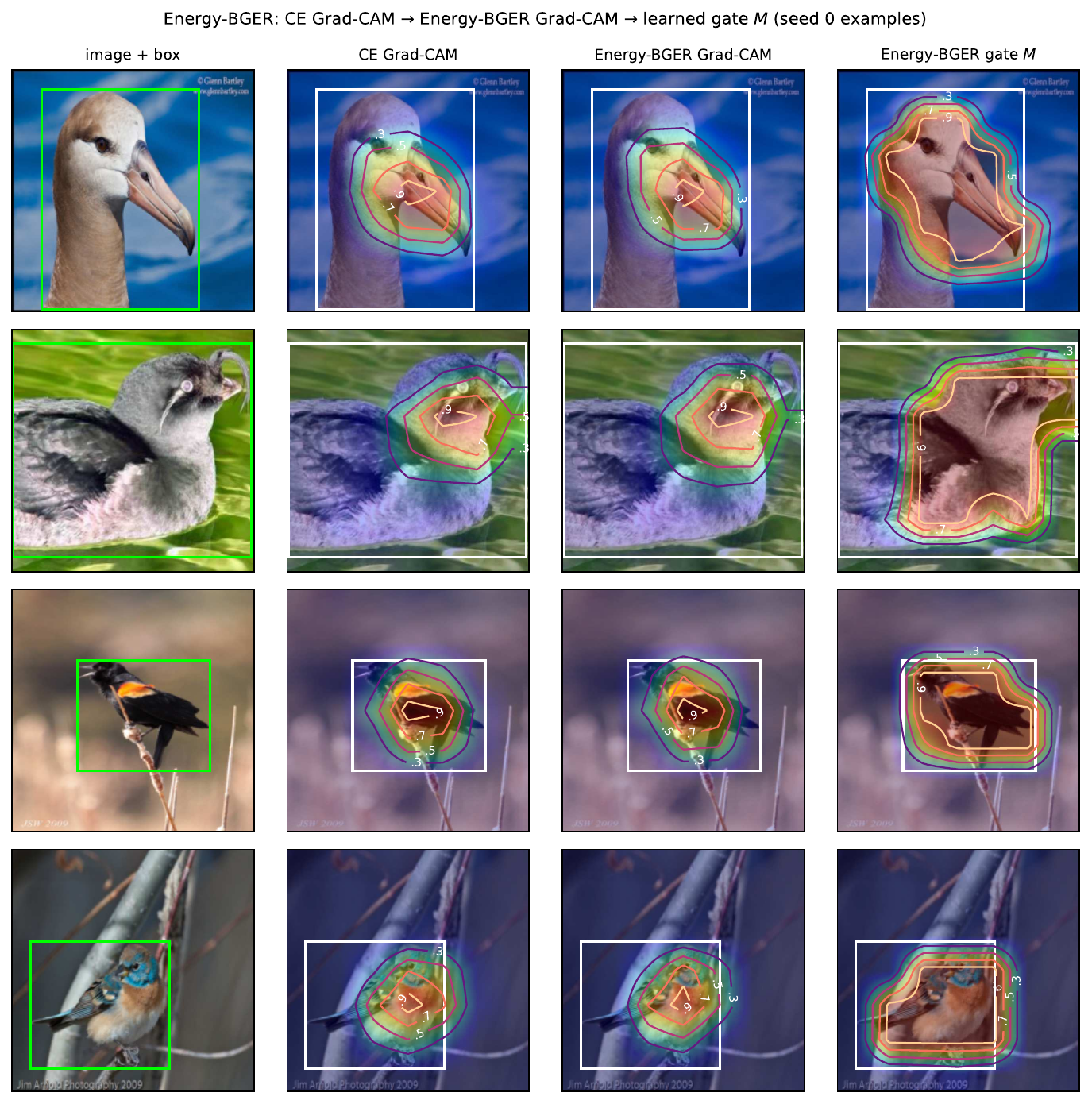}
    \caption{Energy-BGER route-to-CAM examples (seed 0). Columns show the input and GT box, CE and Energy-BGER predicted-class Grad-CAM, and diagnostic gate $M$; contours span $0.3$--$0.9$.}
    \label{fig:gate-transfer}
\end{figure*}

\begin{figure*}[tb]
    \centering
    \includegraphics[width=\textwidth]{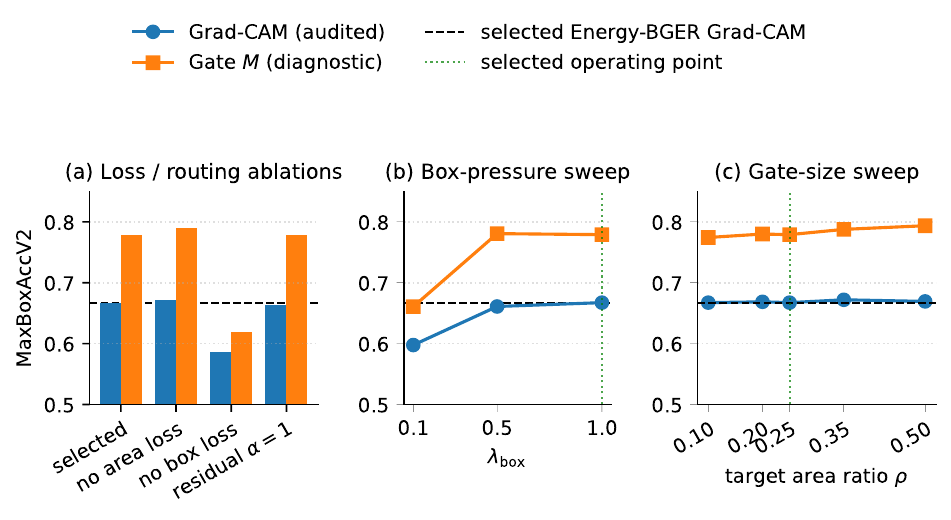}
    \caption{Energy-BGER ablations. Removing the box loss weakens both predicted-class Grad-CAM and gate localization.}
    \label{fig:ablation}
\end{figure*}

\begin{figure*}[tb]
    \centering
    \includegraphics[width=\textwidth]{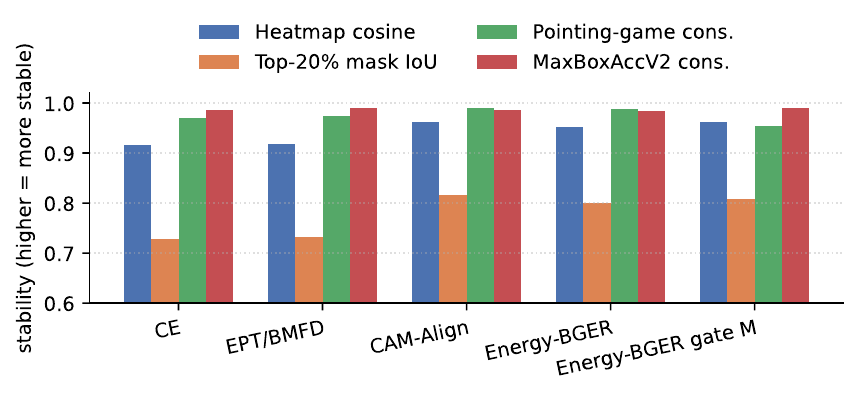}
    \caption{Focus stability under weak augmentations.}
    \label{fig:stability}
\end{figure*}

\begin{figure}[tb]
    \centering
    \includegraphics[width=\columnwidth]{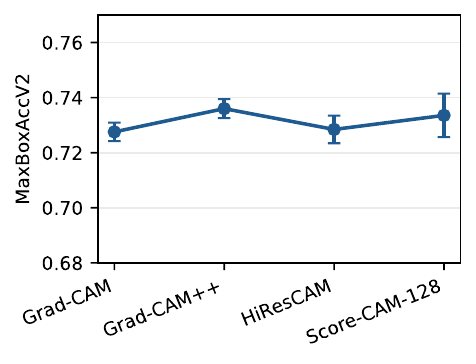}
    \caption{BCE-BGER MaxBoxAccV2 across four CAM operators on the fixed $N=500$ CUB subset (three checkpoints; error bars show seed standard deviation).}
    \label{fig:bce-operator-consistency}
\end{figure}

Figures~\ref{fig:gate-transfer}--\ref{fig:bce-operator-consistency} collect qualitative route-to-CAM examples (Energy recipe), Energy ablation/stability panels, and the BCE-BGER operator-consistency bars corresponding to Tab.~\ref{tab:bce_operator_consistency}.

\section{Attribution-Operator Consistency}

Table~\ref{tab:altcam-fulltest} reports the full-test Energy-BGER comparison; Tab.~\ref{tab:bce_operator_consistency} uses a fixed $N=500$ subset for BCE-BGER.

\paragraph{Attribution-operator consistency.}
CE, BCE-BGER, and CAM-BCE are evaluated with Grad-CAM, Grad-CAM++, HiResCAM, and the gradient-free Score-CAM-128 approximation on identical checkpoints and images (Tab.~\ref{tab:bce_operator_consistency}). BCE-BGER improves over CE on \emph{every} operator (MaxBoxAccV2 $.728$--$.736$ vs.\ CE $.590$--$.602$; EIB $.835$--$.852$ vs.\ $.772$--$.795$), and CAM-BCE sits above both on MaxBoxAccV2 ($.778$--$.804$) consistent with the main-text frontier. The operators share the same box-supervised representation, so this establishes consistency of the CE$\to$route gain across readout rules rather than attribution-method independence.

The class-balanced $N=500$ subset is fixed across operators and seeds, with image IDs recorded in the manifest. Maps are predicted-class conditioned at the pre-gate representation and evaluated at $224^2$; connected components use $112^2$, 21 thresholds on $[0,1]$, and IoU criteria $0.3/0.5/0.7$.

\begin{table}[t]
\centering
\small
\setlength{\tabcolsep}{3pt}
\begin{tabular}{llrr}
\toprule
Operator & Method & EIB $\uparrow$ & MaxBoxV2 $\uparrow$ \\
\midrule
Grad-CAM (main) & CE & $.788{\pm}.001$ & $.584{\pm}.001$ \\
 & CAM-Align & $\mathbf{.866}{\pm}.001$ & $.651{\pm}.005$ \\
 & Energy-BGER & $.859{\pm}.001$ & $\mathbf{.658}{\pm}.001$ \\
\midrule
HiResCAM & CE & $.788{\pm}.001$ & $.584{\pm}.001$ \\
 & CAM-Align & $\mathbf{.866}{\pm}.001$ & $.651{\pm}.004$ \\
 & Energy-BGER & $.863{\pm}.001$ & $\mathbf{.656}{\pm}.001$ \\
\midrule
Grad-CAM++ & CE & $.787{\pm}.001$ & $.592{\pm}.001$ \\
 & CAM-Align & $\mathbf{.866}{\pm}.001$ & $.658{\pm}.005$ \\
 & Energy-BGER & $.849{\pm}.001$ & $\mathbf{.660}{\pm}.001$ \\
\bottomrule
\end{tabular}
\caption{Full-test operator check for Energy-BGER (CUB, $N=5{,}794$, predicted-class maps, ResNet-50, three seeds).}
\label{tab:altcam-fulltest}
\end{table}

\begin{table*}[t]
\centering
\small
\setlength{\tabcolsep}{6pt}
\begin{tabular}{llccc}
\toprule
Method & Operator & EIB $\uparrow$ & PG $\uparrow$ & MaxBoxV2 $\uparrow$ \\
\midrule
CE & Grad-CAM & $.795{\pm}.001$ & $.957{\pm}.005$ & $.593{\pm}.003$ \\
 & Grad-CAM++ & $.794{\pm}.001$ & $.955{\pm}.002$ & $.602{\pm}.001$ \\
 & HiResCAM & $.795{\pm}.001$ & $.957{\pm}.005$ & $.593{\pm}.003$ \\
 & Score-CAM$^*$ & $.772{\pm}.001$ & $.975{\pm}.009$ & $.590{\pm}.001$ \\
\midrule
BCE-BGER & Grad-CAM & $.846{\pm}.002$ & $.991{\pm}.002$ & $.728{\pm}.003$ \\
 & Grad-CAM++ & $.837{\pm}.003$ & $.988{\pm}.002$ & $.736{\pm}.003$ \\
 & HiResCAM & $\mathbf{.852}{\pm}.002$ & $.991{\pm}.002$ & $.728{\pm}.005$ \\
 & Score-CAM$^*$ & $.835{\pm}.002$ & $\mathbf{.995}{\pm}.003$ & $.734{\pm}.008$ \\
\midrule
CAM-BCE & Grad-CAM & $.842{\pm}.012$ & $.994{\pm}.002$ & $.778{\pm}.004$ \\
 & Grad-CAM++ & $.837{\pm}.012$ & $.993{\pm}.002$ & $\mathbf{.804}{\pm}.008$ \\
 & HiResCAM & $.842{\pm}.012$ & $.994{\pm}.002$ & $.778{\pm}.004$ \\
 & Score-CAM$^*$ & $.827{\pm}.012$ & $.995{\pm}.005$ & $.798{\pm}.008$ \\
\bottomrule
\end{tabular}
\caption{Operator consistency on the fixed $N=500$ CUB subset (three seeds, identical images; $^*$Score-CAM-128 approximation). BCE-BGER improves over matched CE on every operator; CAM-BCE is highest on MaxBoxAccV2, consistent with the accuracy--localization frontier.}
\label{tab:bce_operator_consistency}
\end{table*}

\section{Same-Annotation Baselines on Stanford Dogs}

The Dogs topology controls repeat the CUB comparison using the original Energy objective and the same dog box for every supervised row (Tab.~\ref{tab:dogs-baselines}). Energy-BGER reaches $.801$, above the tested input-space and attention controls; the BCE-BGER result is separately reported as $.832$ in the main paper.

Table~\ref{tab:dogs-full} is the full Stanford Dogs replication (referenced from the main-paper Results; full test $N{=}8{,}580$, three seeds). BCE-BGER gives the highest thresholded localization among the tested routing objectives at broadly comparable accuracy.

\begin{table*}[t]
\centering
\begin{minipage}[t]{0.53\textwidth}
\centering
\small
\setlength{\tabcolsep}{2.5pt}
\begin{tabular}{lrrrrr}
\toprule
Method & Acc & ECE15 & EIB & PG & MaxBoxV2 \\
\midrule
CE & $.880{\pm}.005$ & $.045{\pm}.003$ & $.890$ & $.984$ & $.757{\pm}.004$ \\
CAM-Align & $.875{\pm}.002$ & $.041{\pm}.003$ & $.929$ & $.987$ & $.773{\pm}.006$ \\
Energy-BGER & $.877{\pm}.003$ & $.042{\pm}.001$ & $.921$ & $.986$ & $.801{\pm}.002$ \\
Dice-BGER & $.878{\pm}.003$ & $.044{\pm}.001$ & $.908$ & $.993$ & $.829{\pm}.003$ \\
BCE-BGER & $.878{\pm}.003$ & $.046{\pm}.002$ & $.910$ & $.993$ & $\mathbf{.832}{\pm}.000$ \\
\bottomrule
\end{tabular}
\captionof{table}{Stanford Dogs replication (full test $N{=}8{,}580$, three seeds; mean$\pm$std where variance is reported). EIB and PG are three-seed means.}
\label{tab:dogs-full}
\end{minipage}\hfill
\begin{minipage}[t]{0.45\textwidth}
\centering
\small
\setlength{\tabcolsep}{2.5pt}
\begin{tabular}{llrr}
\toprule
Method & Box use & Acc & MaxBoxAccV2 \\
\midrule
CE & none & $.880$ & $.757$ \\
Bg-blackout & input mask & $.872$ & $.771$ \\
Bg-random & input noise & $.865$ & $.765$ \\
Object-crop & input crop & $.847$ & $.715$ \\
ABN-box & supervised attention & $.808$ & $.764$ \\
CAM-Align & CAM proxy & $.875$ & $.773$ \\
Energy-BGER & feature route & $.877$ & $\mathbf{.801}$ \\
\bottomrule
\end{tabular}
\captionof{table}{Same-annotation topology controls for the original Energy objective on Stanford Dogs (three seeds).}
\label{tab:dogs-baselines}
\end{minipage}
\end{table*}

\section{Annotation-Fraction Curve}
\label{sec:fraction-appendix}

Energy-BGER is trained with nested class-stratified box fractions and no per-fraction tuning. At $10\%$ ($400$ images), MaxBoxAccV2 reaches $.631$ versus CE $.588$ and full-budget Energy-BGER $.663$.

\begin{table}[t]
\centering
\footnotesize
\setlength{\tabcolsep}{1pt}
\begin{tabular}{rrrrrrr}
\toprule
Boxes & $n$ & Acc & ECE15 & EIB & MaxBoxV2 & HCU \\
\midrule
10\% & 400 & $.749{\pm}.001$ & $.020{\pm}.002$ & $.836{\pm}.002$ & $.630{\pm}.003$ & $.19\%$ \\
25\% & 1200 & $.751{\pm}.003$ & $.019{\pm}.004$ & $.848{\pm}.001$ & $.646{\pm}.003$ & $.15\%$ \\
50\% & 2400 & $.747{\pm}.002$ & $.019{\pm}.003$ & $.854{\pm}.001$ & $.652{\pm}.000$ & $.07\%$ \\
100\% & 4794 & $.751{\pm}.001$ & $.017{\pm}.001$ & $.861{\pm}.001$ & $.663{\pm}.004$ & $.04\%$ \\
\bottomrule
\end{tabular}
\caption{Annotation-fraction curve (CUB, ResNet-50): the box-routing loss is applied to a class-stratified fraction of training images (nested subsets); all images keep class labels. Spatial metrics on a fixed 2{,}500-image test subset; \textsuperscript{d}\,=\,seed-0 diagnostic, others three seeds. CE reference (no boxes): MaxBoxAccV2 $.588$, EIB $.790$, HCU $1.10\%$. Ten percent of boxes already recovers most of the transfer; gains grow monotonically with budget at CE-level accuracy and ECE.}
\label{tab:fraction-curve}
\end{table}

\begin{figure}[tb]
    \centering
    \includegraphics[width=\columnwidth]{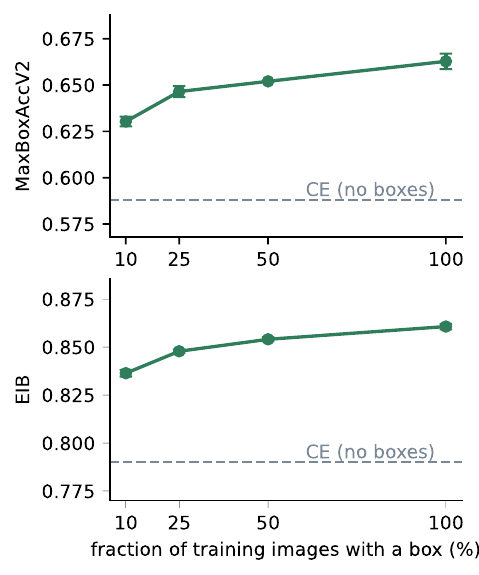}
    \caption{Energy-BGER route-to-CAM transfer versus CUB training-box fraction. The dashed line is CE; error bars show seed standard deviation where three seeds are available.}
    \label{fig:fraction-curve}
\end{figure}

\section{External CUB WSOL References}
\label{app:recent}

Table~\ref{tab:external-wsol} lists dedicated WSOL systems evaluated locally under our image/box protocol; they output true-class localization maps and are references, not classifier predicted-class Grad-CAM baselines. The three systems are the Spatial-Aware Token method (SAT; \citealp{wu2023sat}), the Category-aware Allocation TRansformer (CATR; \citealp{chen2023catr}), and the Token Semantic Coupled Attention Map method (TS-CAM; \citealp{gao2021tscam}). Each was evaluated with its official public code and its official released CUB checkpoint (cloned and downloaded locally; the released audit records list the exact repository and checkpoint file per method).

\begin{table*}[tb]
\centering
\small
\begin{tabular}{lllrrrr}
\toprule
Method & Supervision & Map & Acc & EIB & PG & MaxBoxAccV2 \\
\midrule
SAT \citep{wu2023sat} & image labels & WSOL map & .815 & .628 & .901 & .923 \\
CATR \citep{chen2023catr} & image labels & WSOL attn. & .842 & .607 & .865 & .783 \\
TS-CAM \citep{gao2021tscam} & image labels & WSOL CAM & .729 & .730 & .991 & .781 \\
\bottomrule
\end{tabular}
\caption{External CUB WSOL references (single official checkpoint each), evaluated locally under our image/box protocol. These output dedicated true-class WSOL maps and are \emph{not} comparable to the classifier's predicted-class Grad-CAM in Tab.~\ref{tab:bger-cub-external}; listed as localization references. ECE15 is NA (calibration not reliably exported).}
\label{tab:external-wsol}
\end{table*}

\section{External WSOL Protocol Check}
\label{app:external-protocol}

Before adding SAT, CATR, and TS-CAM to the CUB comparison, we audited whether their local scoremaps were compatible with the same CUB protocol used for BGER. All three methods pass the hard checks: exact CUB test-set match (5,794/5,794 images), zero train overlap, zero missing/extra/duplicate image identifiers, zero bbox mismatches, zero scoremap shape mismatches, zero missing/non-finite scoremaps, and the same dense 0.001 threshold sweep for MaxBoxAccV2.

We also spot-checked alignment visually, overlaying each method's locally exported true-class scoremap and the GT box on twelve random CUB test images per method; all confirmed correct image/box/scoremap registration (overlays omitted for brevity). The audit also clarifies the comparison boundary. SAT, CATR, and TS-CAM use true-class dedicated WSOL maps. Their localization maps are not the same object as BGER's post-hoc classifier Grad-CAM, and they should not be presented as direct substitutes for BGER. They are included as external localization references with explicit supervision, map type, map conditioning, and protocol-decision columns in Tab.~\ref{tab:bger-cub-external}.

\section{Readout Consistency and RSNA Subgroups}
\label{app:readout-geometry}
These two tables are referenced from the main-paper Results; they are placed here for space.

\begin{table*}[t]
\centering
\begin{minipage}[t]{0.44\textwidth}
\centering
\small
\setlength{\tabcolsep}{4pt}
\begin{tabular}{lrrr}
\toprule
Readout & CE & BCE-BGER & \textit{CAM-BCE} \\
\midrule
Grad-CAM & $.593$ & $.728$ & \textit{.778} \\
Grad-CAM++ & $.602$ & $.736$ & \textit{.804} \\
HiResCAM & $.593$ & $.728$ & \textit{.778} \\
Score-CAM-128 & $.590$ & $.734$ & \textit{.798} \\
\bottomrule
\end{tabular}
\captionof{table}{MaxBoxAccV2 across CAM readouts (CUB, fixed $N{=}500$ subset, three seeds, identical images and checkpoints). Italics denote direct CAM-proxy supervision. Full metrics: App.~Tab.~\ref{tab:bce_operator_consistency}.}
\label{tab:operators}
\end{minipage}\hfill
\begin{minipage}[t]{0.52\textwidth}
\centering
\small
\setlength{\tabcolsep}{4pt}
\begin{tabular}{lrrr}
\toprule
Subgroup & $n$ & $\Delta$EIB & $\Delta$BestIoU \\
\midrule
all positive boxed & $1422$ & $+.200$ & $+.167$ \\
\midrule
single-box & $643$ & $+.131$ & $+.113$ \\
multi-box ($\ge 2$) & $779$ & $+.257$ & $+.212$ \\
\midrule
small lesions & $474$ & $+.088$ & $+.098$ \\
medium lesions & $474$ & $+.194$ & $+.205$ \\
large lesions & $474$ & $+.319$ & $+.199$ \\
\midrule
EIB up, IoU down & $181$ & $+.133$ & $-.118$ \\
\bottomrule
\end{tabular}
\captionof{table}{RSNA per-case CE$\to$BCE-BGER change by target geometry (positive boxed test cases, seed 0); the final row is not a geometry subgroup but reports the frequency and mean magnitude of metric-discordant cases. Soft-energy gain grows with lesion size and box count; box-overlap gain plateaus. Case-by-case, CAM-BCE's soft-energy advantage over BCE-BGER is nearly constant ($-.199$ overall, $|\Delta|\le.04$ across subgroups) while their box overlap is indistinguishable ($+.001$, $|\Delta|\le.005$), so the route-vs-proxy ordering is a metric property, not a geometry effect. BestIoU is the maximum per-image box IoU across thresholds; it is used for subgroup analysis and is distinct from the dataset-level MaxBoxAccV2 in Tab.~\ref{tab:ext-datasets}. Area terciles at $.052$/$.124$ of the image. Controlled extension, not clinical validation.}
\label{tab:rsna-subgroup-bce}
\end{minipage}
\end{table*}

\end{document}